\documentclass[lettersize,journal]{IEEEtran}
\usepackage{amsmath,amsfonts}
\usepackage{array}
\usepackage[caption=false,font=normalsize,labelfont=sf,textfont=sf]{subfig}
\usepackage{textcomp}
\usepackage{stfloats}
\usepackage{url}
\usepackage{verbatim}
\usepackage{graphicx}
\usepackage{cite}
\usepackage{tikz}
\usetikzlibrary{spy}
\usepackage{multirow}
\usepackage{xspace}
\usepackage{caption}
\usepackage[ruled,vlined]{algorithm2e}
\usepackage[pagebackref,breaklinks,colorlinks]{hyperref}
\usepackage[capitalize]{cleveref}
\usepackage{tabularx}

\newcommand{\myfore}[0]{NSC}
\newcommand{\myback}[0]{SEID\xspace}
\newcommand{\mydata}[0]{NED\xspace}

\newcommand{\ie}{{\emph{i.e.}}\xspace}

\newcommand{\eg}{{\emph{e.g.}}\xspace}
\newcommand{\etal}{{\emph{et al.}}\xspace}

\begin{document}

\title{Non-Uniform Exposure Imaging via Neuromorphic Shutter Control}

\author{Mingyuan Lin, Jian Liu, Chi Zhang, Zibo Zhao, Chu He, and Lei Yu
\thanks{M. Lin, J. Liu, C. Zhang, Z. Zhao, C. He, and L. Yu are with the School of Electronic Information, Wuhan University, Wuhan 430072, China. E-mail: \{linmingyuan, jianliu1101, zhangchi1, zibozhao, chu.he, ly.wd\}@whu.edu.cn.}
\thanks{The research was partially supported by the National Natural Science Foundation of China under Grants 62271354 and 41371342.}
\thanks{Corresponding authors: C. He and L. Yu.}}

\markboth{Submission to IEEE}%
{Shell \MakeLowercase{\textit{et al.}}: A Sample Article Using IEEEtran.cls for IEEE Journals}


\maketitle

\begin{figure*}
    \centering
    \includegraphics[width=\textwidth]{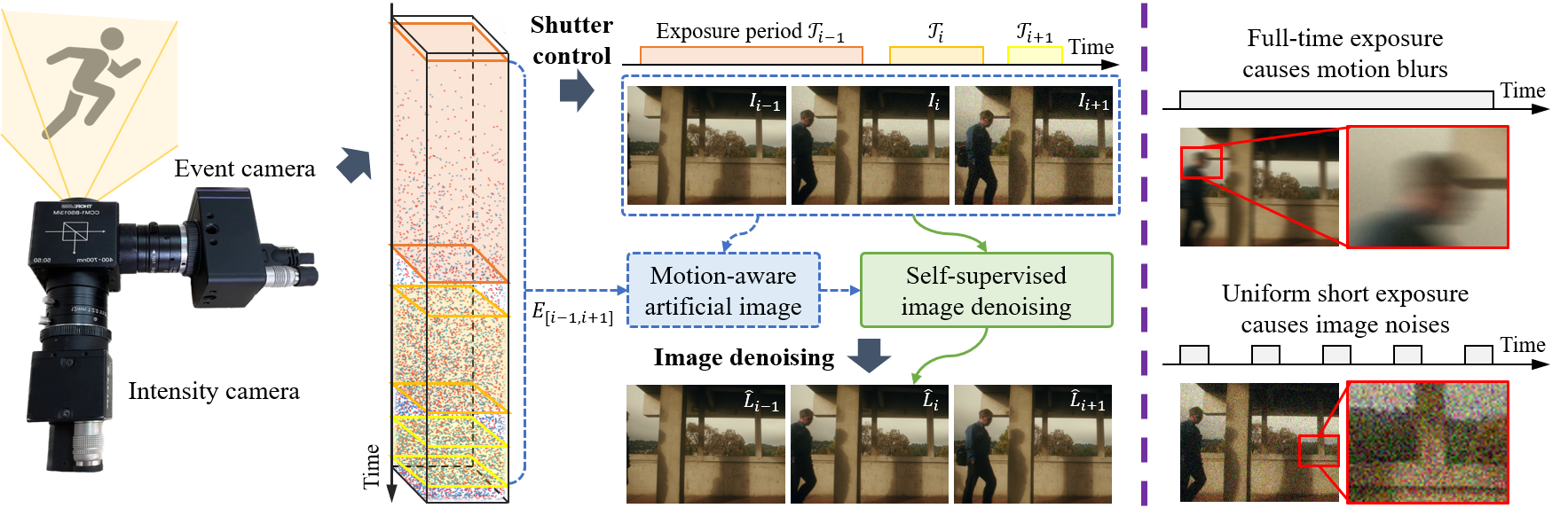}
    \caption{Comparisons of our neuromorphic exposure imaging system (left) with the conventional exposure setups (right). Our system contains two parts, \ie, Neuromorphic Shutter Control (NSC) utilizing the event camera to monitor the scene motion information and adjust the camera shutter in real time, and Self-supervised Event-based Image Denoising (\myback) stabilizing high SNR for each frame. Green solid lines indicate the process of image denoising and blue dashed lines indicate the generation of the self-supervised signals in \myback.}
    \label{fig:first}
\end{figure*}

\begin{abstract}
By leveraging the blur-noise trade-off, imaging with non-uniform exposures largely extends the image acquisition flexibility in harsh environments. However, the limitation of conventional cameras in perceiving intra-frame dynamic information prevents existing methods from being implemented in the real-world frame acquisition for real-time adaptive camera shutter control. To address this challenge, we propose a novel Neuromorphic Shutter Control (\myfore) system to avoid motion blurs and alleviate instant noises, where the extremely low latency of events is leveraged to monitor the real-time motion and facilitate the scene-adaptive exposure. Furthermore, to stabilize the inconsistent Signal-to-Noise Ratio (SNR) caused by the non-uniform exposure times, we propose an event-based image denoising network within a self-supervised learning paradigm, \ie, \myback, exploring the statistics of image noises and inter-frame motion information of events to obtain artificial supervision signals for high-quality imaging in real-world scenes. To illustrate the effectiveness of the proposed \myfore, we implement it in hardware by building a hybrid-camera imaging prototype system, with which we collect a real-world dataset containing well-synchronized frames and events in diverse scenarios with different target scenes and motion patterns. Experiments on the synthetic and real-world datasets demonstrate the superiority of our method over state-of-the-art approaches. 


\end{abstract}

\begin{IEEEkeywords}
Non-uniform exposure imaging, shutter control, event camera, self-supervised learning.
\end{IEEEkeywords}

\section{Introduction}
\IEEEPARstart{I}{n} modern photography, capturing high-quality images heavily depends on the control of the camera exposure. Imaging with long exposure time performs well in static scenes while causing significant motion blurs in dynamic scenes \cite{jin2018learning,cho2021rethinking,zhang2021exposure,song2022cir,zhou2023deblurring}. Although shortening the exposure time helps to avoid blurs, another type of distortion, the notorious noises, starts to dominate \cite{mildenhall2018burst,mustaniemi2018lsd,xia2020basis,izadi2023image}. In reality, the random variation between dynamic and static scenes (\eg, the person entering the camera view as shown in \cref{fig:first}) prevents uniform exposure imaging from maintaining high-quality and stable imaging. 

Instead, the non-uniform exposure imaging strategy controls the camera shutter in response to the scene variations for better imaging \cite{dahary2021digital,chang2021low,zhao2022d2hnet,zhang2023deep,han2023camera}. This strategy extends the exposure time to alleviate noises for static scenes and shortens the exposure time to avoid motion blurs in dynamic scenes and has demonstrated its superiority in non-linear dynamic scenes compared to uniform exposure imaging. However, conventional cameras are limited in perceiving and processing the intra-frame dynamic information during the exposure. Therefore, existing solutions either adopt an offline post-processing scheme employed for a captured video \cite{dahary2021digital,chang2021low,zhao2022d2hnet}, sacrifice the robustness of the algorithm by fixing a predefined exposure setting for various scenarios \cite{zhang2023deep}, or rely on the assumption of the uniform and linear motion, which is not always held in real-world scenarios \cite{han2023camera}. Therefore, a challenging problem arises: \textit{Can we break the intra-frame scene motion unawareness and control the camera shuttle in real time?}


In this paper, we show that the above-mentioned problem can be effectively solved via the proposed use of the event camera to assist with shutter control. The event camera is a neuromorphic sensor that encodes the light intensity variations and asynchronously reports event data with extremely low latency \cite{lichtsteiner128Times1282008,gallego2020event}. Thus, we propose a Neuromorphic Shutter Control (\myfore) system, explicitly intervening in the acquisition process of the conventional cameras in a real-time manner with the assistance of the event camera. Since it is proved that recovering the high-quality result from a noisy image has more potential than from a blurry one \cite{zhang2022exploring}, our \myfore\ is designed to tend to shorten the exposure time to capture motion blur-free image in the high dynamic scenes. Specifically, we propose multiple effective strategies to Measure real-time Motion by leveraging the statistics of Events (EMM) without requiring time-consuming operations, \eg, event stacking. We implement the proposed \myfore\ in hardware and experiment with it under diverse conditions, \eg, various motion speeds and illuminations, validating the real-time performance and robustness of our \myfore.

Due to the adaptive extension or shortening of exposure time in the non-uniform exposure imaging, the captured frames exhibit inconsistent Signal-to-Noise Ratio (SNR) in temporal dimension \cite{barakat2008tradeoff}. Alternative back-end image denoising methods have been proposed recently to stabilize the SNR and achieve satisfactory high-quality images \cite{tassano2020fastdvdnet,vaksman2021patch,zamir2021multi}. Furthermore, to overcome the difficulty in collecting the paired noisy and ground-truth clean images in real-world scenarios, several methods assume independent and zero-mean noises, and employ the self-supervised learning framework, \ie, utilizing the consecutive frames to construct the artificial supervision signals \cite{vaksman2023patch}. However, the dynamic scenes can cause the misplacement of textures across neighboring frames, rendering the supervision signals unreliable and resulting in blurring of the original image textures. 


Therefore, we propose a Self-supervised Event-based Image Denoising (\myback) framework, purposing to enhance the frames captured with our \myfore\ for the high-quality imaging. By considering the zero-mean characteristics of noise, we can filter it into a clean one by integrating with neighboring frames, where the inter-frame motion information from event data can guide us to compensate for the misalignment of neighboring frames \cite{tulyakov2021time}. For training, we adopt neighboring noisy shots and inter-frame event streams to create the artificial target image as the supervision signals. Specifically, since the event data continuously records the motion of textures, it helps to eliminate the moving textures while retaining static textures and content for the reliable supervision.

The main contributions of this paper are three-fold:
\begin{itemize}
    \item We propose a Neuromorphic Shutter Control (\myfore) system. To the best of our knowledge, it is the first system to proactively monitor the real-time scene motion information and explicitly intervene in the shutter controlling and frame acquisition process.
    \item We propose a Self-supervised Event-based Image Denoising (\myback) framework, integrating neighboring frames as the target images and leveraging events to avoid interference from unreliable blurry regions.
    \item We build a hybrid prototype system to illustrate the efficiency of our \myfore\ and collect a real-world dataset, named Neuromorphic Exposure Dataset (\mydata), containing frames captured with \myfore\ and corresponding events in various scenes and motion patterns. 
\end{itemize}

\section{Related Work}
\subsection{Exposure Controlled Imaging}
It is well known that recovering sharp and clean results from captured frames depends heavily on not only the ability of restoration methods but also the quality of input images. Many methods explore controlling the exposure time of input images to obtain better reconstruction results \cite{dahary2021digital,chang2021low,zhang2022exploring,zhang2023deep}. Zhang \etal, \cite{zhang2022exploring} propose the concept of image restoration potential (IRP) to filter valuable frames for the restoration process. Chang \etal, \cite{chang2021low} feed their image fusion network with the long- and short-exposure frame pairs for the low-light image enhancement task. Furthermore, Zhang \etal, \cite{zhang2023deep} and Dahary \etal, \cite{dahary2021digital} attempt to directly encode the camera shutter settings for better imaging, both proposing end-to-end frameworks to co-optimize shutter controller and image restoration modules jointly. The former interests in the deblurring task and encodes the flutter shutter to capture a blurry image with tailored invertible blur kernels, while the latter jointly considers the burst denoising and deblurring task. However, due to the limitation that conventional cameras cannot perceive scene motion information while exposing, the adaptability of the above methods to complex and dynamic real-world scenes is inevitably hindered.

\subsection{Self-supervised Image Denoising}
Relying on the assumption that noises are pixel-wise, independent, and zero-mean, recent researchers focus on the single-frame denoising with the self-supervised learning framework \cite{lehtinen2018noise2noise,krull2019noise2void,laine2019high,huang2021neighbor2neighbor,lee2022ap,li2023spatially}. Among them, Blind-Spot Network (BSN) is a notable method that directly treats the input noisy image as the target image and reconstructs a clear pixel from the surrounding noisy pixels, independent of the corresponding input pixel \cite{krull2019noise2void}. Based on BSN, several approaches have achieved better performance via various strategies, \ie, neighbor sub-sampling \cite{huang2021neighbor2neighbor}, asymmetric pixel-shuffle downsampling \cite{lee2022ap}, and flat-texture separated training\cite{li2023spatially}. To avoid producing over-smooth results due to the similarity between high-frequency noises and the scene textures as the above methods, Vaksman \etal, \cite{vaksman2023patch} incorporate the temporal consistency of noisy videos and perform better. They utilize patch matching and stitching techniques to construct artificial patch-craft images from the neighboring frames of the input noisy frame. These crafted images are then used as training targets for the algorithm. However, it heavily relies on the similarity between frames and is susceptible to patch misplacement caused by significant inter-frame motion.

\subsection{Event-aided Imaging}
The event camera, as a bio-inspired vision sensor, poses a paradigm shift in visual information acquisition \cite{lichtsteiner128Times1282008,gallego2020event}. Differing from the traditional cameras which capture intensity images at a fixed frame rate, event cameras only respond to the brightness change and emit asynchronous events composed of pixel position, timestamp, and polarity. This mechanism leads to many promising properties, \eg, low latency and high temporal resolution \cite{gallego2020event}, and is beneficial for high-quality imaging tasks, especially under harsh conditions such as high-speed motion. For the motion deblurring task, the microsecond-level high temporal resolution of events enables continuous recording of dynamic scenes, especially the high-contrast edges, and helps to alleviate the motion ambiguity and compensate for the erased textures in the blurry image \cite{pan2019bringing,wang2020event,xu2021motion,sun2022event,zhang2023generalizing}. With the assistance of the inter-frame information that events offer, recent event-based frame interpolation methods can recover accurate intermediate frames even under non-linear motions \cite{lin2020learning,tulyakov2021time,tulyakov2022time,zhang2022unifying}. However, current event-aided imaging methods primarily concentrate on the back-end image processing step, heavily relying on the quality of captured images. Hence, they fail to fully harness the potential of the event camera for the real-time high-quality imaging task in dynamic scenes.

\section{Problem Formulation}
With a predefined exposure length $T$, the modern imaging of conventional cameras can be described as,
\begin{equation}
    L = \mathcal{P}(\mathcal{A}(I(t), T)),
\end{equation}
where $L$ is the target clear and sharp image, $\mathcal{A}(\cdot)$ donates the frame acquisition process of cameras, and $\mathcal{P}(\cdot)$ is the digital image enhancement function \cite{battiato2010image}. The conventional camera continuously collects photons that reach the sensor and generates electrons when the shutter is open. Hence, the captured frame can be formulated as the integral of the latent frames during the exposure time $t\in[0,T]$ as,
\begin{equation}
    \mathcal{A}(I(t), T)=\int^T_{t=0} I(t)dt,
    \label{eq:f_it}
\end{equation}
where $I(t)$ is the instant intensity of pixels at time $t$. In particular, the instant intensity is degraded by noise with the Gaussian-Poisson distribution, which can be modeled as a signal-dependent Gaussian distribution \cite{healey1994radiometric,xia2020basis,izadi2023image},
\begin{equation}
    I(t)\sim\mathcal{N}(L(t), \sigma_pL(t)+\sigma^2_g),
    \label{eq:i_l+n}
\end{equation}
where $L(t)$ denotes the noise-free intensity of pixels. $\mathcal{N}$ is the Gaussian distribution, and $\sigma_p$ and $\sigma_g$ are the noise parameters. With \cref{eq:f_it,eq:i_l+n}, we can infer that in dynamic scenes, $T$ plays a crucial role in determining the quality of the captured frames, that is, a frame with a long/short exposure time tends to be blurry/noisy.

Hence, we propose the shutter control task for non-uniform imaging to utilize real-time motion information from the scene as a direct reference to control each specific shutter switch,
\begin{equation}
    L = \mathcal{P}(\mathcal{A}(I(t), \mathcal{M}(t))),
    \label{eq:f_im}
\end{equation}
where $\mathcal{M}(t)$ means the scene Motion Measure (MM) function, whose definition is the key point for the real-time shutter control. The definition of $\mathcal{A}(\cdot)$ can be rewritten as,
\begin{equation}
        \mathcal{A}(I(t), \mathcal{M}(t)) \doteq \left\{\begin{array}{ll}
        0, & \text { if } t < 0 \text{ or } \mathcal{M}(t) > R, \\
        \int^t_0 I(t)dt, & \text { if } t \ge 0 \text{ and } \mathcal{M}(t) \le R,
        \end{array}\right.
    \label{eq:m}
\end{equation}
where $R$ is a threshold value referring to the maximum magnitude of motion that can be accommodated in a high-quality image to avoid blurs. 

Further, to handle the temporal inconsistency SNR of captured frames due to the varied exposure time, $\mathcal{P}(\cdot)$ is defined as an image denoising function. Due to the difficulty in capturing the paired real noisy and clean images for the learning of $\mathcal{P}(\cdot)$, developing a self-supervised learning framework is required to get rid of the dependence on ground-truth images.

In a word, to realize the motion-aware non-uniform exposure imaging defined in \cref{eq:f_im}, challenges still exist.
\begin{itemize}
    \item Since the exposure time $T$ of cameras is typically in the order of $ms$, the data captured to calculate the scene motion requires extremely high temporal resolution.
    \item The target MM function $\mathcal{M}(t)$, which transforms the data to motion information, should have very low computational complexity to avoid heavy time consumption.
    \item While recent methods have realized self-supervised learning for image denoising, they inevitably compromise the original textures of images, as their similarity to high-frequency noises.
\end{itemize}

\section{Method}
\subsection{Preliminaries}
Unlike conventional cameras, each pixel of an event camera responds to changes in the radiance $L(t)$ and generates streams of asynchronous events \cite{lichtsteiner128Times1282008,gallego2020event}. For an event camera, the $i$-th event $e_i=(\mathbf{x}_i, t_i, p_i)$ is triggered at position $\mathbf{x}_i=(x_i,y_i)^T$ and time $t_i$ whenever the logarithmic pixel intensity difference reaches a threshold $C$,
\begin{equation}
    \tilde{L}(t_i,\mathbf{x}_i) - \tilde{L}(t_i-\Delta t_i,\mathbf{x}_i) = p_iC,
    \label{eq:ll_pc}
\end{equation}
where $\tilde{L}(\cdot,\mathbf{x})$ corresponds to the logarithmic intensity of pixel $\mathbf{x}$, $\Delta t_i$ is the time difference between the current and the previous event, and $p\in\{+1,-1\}$ denotes the polarity showing the direction of brightness change. Thus, the increment in radiance within a time window $\Delta t$ is expressed as the accumulation of the event data,
\begin{equation}
    \Delta\tilde{L}(\mathbf{x}, t) = \sum_{e_i\in \mathcal{E}_{[t-\Delta t,t]}} p_iC,
    \label{eq:dl_pc}
\end{equation}
where $\mathcal{E}_{[t-\Delta t,t]}$ denotes the event streams emitted within the time period $[t-\Delta t,t]$.

\begin{figure}[t]
    \centering
    \includegraphics[width=0.975\linewidth]{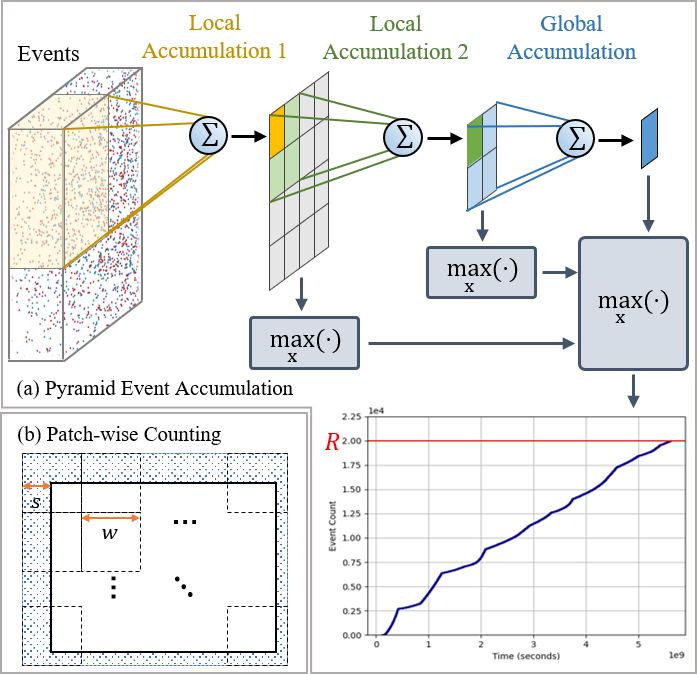}
    \caption{(a) Pyramid Event Accumulation (PEA) is designed for the local motion. (b) Demonstration of the patch-wise counting for PEA with the parameters $s$ and $w$. $R$ is a threshold value defined in \cref{eq:m}.}
    \label{fig:pea}
\end{figure}

\subsection{Neuromorphic Shutter Control (\myfore)}
\label{sec:nsc}
\subsubsection{Optical Flow for Motion Measure}
It is well known that optical flow is a visual expression of motion \cite{horn1981determining}. Thus, the scene Motion Measure (MM) function can be reformulated as an optical flow estimation task as,
\begin{equation}
    \mathcal{M}(t) \doteq \int^t_0 \sum_{\mathbf{x}\in\Omega} |\mathbf{u}(\mathbf{x},t)| dt,
    \label{eq:m_u}
\end{equation}
where $\mathbf{u}(\mathbf{x},t)$ is the optical flow at the pixel position $\mathbf{x}$. The conventional camera lacks the capability to sense motion during the exposure, necessitating the introduction of an additional sensor, \eg, a high-speed intensity camera or an event camera\footnote{Though Inertial Measurement Unit (IMU) is one of the options, it can only provide the ego-motion of the camera but not the motion of targets.}. Based on the data type, we can define two scene motion measure methods, \ie, Frame-based Motion Measure (FMM) and Event-based Motion Measure (EMM). 

For FMM, it takes two consecutive frames with a small $\Delta t$ interval as inputs to predict scene motion by,
\begin{equation}
    \mathbf{u}(t)=\mathcal{F}_{im}(L(t-\Delta t), L(t)),
\end{equation}
where $\mathcal{F}_{im}(\cdot)$ is the frame-based optical flow estimation function. However, the need for the much smaller $\Delta t$ than $T$ to estimate motion during the exposure time $[0,T]$ requires the additional camera to have an extremely high temporal resolution, which however introduces noise. Meanwhile, The time delay incurred from the optical flow computation process cannot be disregarded. 

Similarly, existing event-based optical flow estimation methods \cite{gehrig2021raft,brebion2021real,shiba2022secrets} can be utilized for EMM by,
\begin{equation}
    \mathbf{u}(t)=\mathcal{F}_{ev}(\mathcal{E}_{[t-\Delta t,t]}),
\end{equation}
where $\mathcal{F}_{ev}(\cdot)$ is the event-based optical flow estimation function, and $\mathcal{E}_{[t-\Delta t,t]}$ denotes the event streams emitted within the time period $[t-\Delta t,t]$. We define the shutter control system with the assistance of the event camera and the above EMM function as the Neuromorphic Shutter Control (\myfore) system. However, as classical event-based methods for estimating optical flow need the event compression process from streams to the frame-like format and spatial convolution \cite{brebion2021real}, they are time-consuming for the real-time \myfore.

\subsubsection{Global Event Accumulation}
To address the above issues, we propose Global Event Accumulation (GEA) to use the number of events as the EMM. As in \cite{zhang2022formulating}, under the assumption of the constant illumination, the mutual relationship among the scene textures, radiance difference, and optical flow within the small time period $\Delta t$ can be modeled by,
\begin{equation}
    \Delta\tilde{L}(\mathbf{x},t) \approx -\nabla\tilde{L}(\mathbf{x},t)\cdot \bar{\mathbf{u}}(\mathbf{x})\Delta t,
    \label{eq:iwe}
\end{equation}
determining that events are caused by the moving spatial gradient, $\nabla\tilde{L}=(\delta_x\tilde{L}, \delta_y\tilde{L})^T$, whose motion can be denoted by the optical flow $\bar{\mathbf{u}}(\mathbf{x})\Delta t$ during $\Delta t$. $\bar{\mathbf{u}}(\mathbf{x})$ represents the unit-time optical flow as $\bar{\mathbf{u}}(\mathbf{x}) = \frac{\mathbf{u}(\mathbf{x})}{\Delta t}$. We can convert \cref{eq:iwe} with $\mathbf{u}(\mathbf{x},t)=\bar{\mathbf{u}}(\mathbf{x})\Delta t$ as,
\begin{equation}
    \Delta\tilde{L}(\mathbf{x},t) \approx -\nabla\tilde{L}(\mathbf{x},t)\cdot \mathbf{u}(\mathbf{x},t).
\end{equation}

Over a short time period, we assume that the scene gradient $\nabla\tilde{L}(\mathbf{x},t)$ is constant. Thus, at the position $\mathbf{x}$, the magnitude of motion is proportional to the brightness variation,
\begin{equation}
    |\mathbf{u}(\mathbf{x},t)| \propto |\Delta\tilde{L}(\mathbf{x},t)|.
    \label{eq:u_dl}
\end{equation}

Further, we combine \cref{eq:dl_pc,eq:m_u,eq:u_dl} to propose GEA, redefining EMM with the number of events using Dirac-delta functions $\delta(\cdot)$ as,
\begin{equation}
    \mathcal{M}_\textit{g}(t) \doteq \int^t_0 \sum_{i}\delta(t-t_i)\delta(\mathbf{x}-\mathbf{x}_i) dt.
    \label{eq:m_g}
\end{equation}

Since the event data can be sliced and packaged according to a fixed size number of events and transmitted to the computational devices, \myfore\ with GEA can be implemented by setting the size of each slicer of the event stream without any event-to-frame transformation. In this way, even the data unpacking operation is not needed, leading to a highly time-efficient EMM strategy.

\def\ssxxsone{(-0.1,0.7)} 
\def\ssyysone{(-0.05,-1.8)} 
\def\ssxxstwo{(-0.39,-0.35)} 
\def\ssyystwo{(0.05,-1.8)} 
\def\ssmag{4}
\def\ssizz{1.35cm} 
\def\sswidth{0.32\linewidth} 
\begin{figure}[t]
    \centering
    \begin{tabular}{c c c}
        \hspace{-2mm}
        \begin{tikzpicture}[spy using outlines={green,magnification=\ssmag,size=\ssizz},inner sep=0]
    		\node {\includegraphics[width=\sswidth]{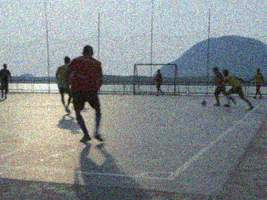}};
    		\spy on \ssxxsone in node [left] at \ssyysone;
            \spy [red] on \ssxxstwo in node [right,red] at \ssyystwo;
    	\end{tikzpicture} & \hspace{-4.8mm}
        \begin{tikzpicture}[spy using outlines={green,magnification=\ssmag,size=\ssizz},inner sep=0]
    		\node {\includegraphics[width=\sswidth]{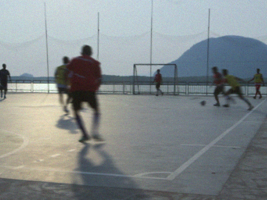}};
    		\spy on \ssxxsone in node [left] at \ssyysone;
            \spy [red] on \ssxxstwo in node [right,red] at \ssyystwo;
    	\end{tikzpicture} & \hspace{-4.8mm}
        \begin{tikzpicture}[spy using outlines={green,magnification=\ssmag,size=\ssizz},inner sep=0]
    		\node {\includegraphics[width=\sswidth]{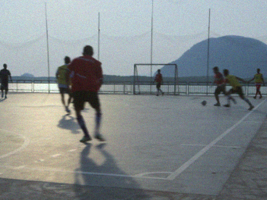}};
    		\spy on \ssxxsone in node [left] at \ssyysone;
            \spy [red] on \ssxxstwo in node [right,red] at \ssyystwo;
    	\end{tikzpicture} \\
        \hspace{-3mm}(a) Instant frame & \hspace{-3mm}(b) GEA result & \hspace{-3mm}(c) PEA result
    \end{tabular}
    \caption{Comparison of the \myfore\ results with two strategies sharing a same threshold $R$. (a) The instant frame with the local motion scene is disturbed by severe noise. (b) \myfore\ with the GEA can significantly suppress noise, while the local blur may arise. (c) \myfore\ with the PEA enhances the sensitivity to the local motion and achieves less severe blurs.}  
    \label{fig:gea_vs_pea}
\end{figure}

\subsubsection{Pyramid Event Accumulation}
With the global event accumulation defined by \cref{eq:m_g}, we can realize the real-time motion-aware shutter control defined by \cref{eq:f_im}. In this section, we delve into the influence of various motion types on camera exposure.

The scene recorded by the captured frame can be divided into two parts: texture representing high-frequency signals and content representing low-frequency signals, where the motion of textures in dynamic scenes is the key cause of blur. Based on the proportion of the blurry area in the image, we can classify the image blur into two categories, \ie, the global blur and the local blur. Specifically, camera or scene motion in scenes with sufficient texture results in the global motion in the frame, while sparse texture or object motion leads to the local blur. Regardless of the location of events, we have realized the monitoring of the global blur by defining GEA in \cref{eq:m_g}. To enhance the sensitivity of the EMM function to the local motion, we further propose Pyramid Event Accumulation (PEA) with patch-wise counting for events, as shown in \cref{fig:pea}.

\begin{figure*}[t]
    \centering
    \begin{tabular}{c c}
    \includegraphics[width=0.3\textwidth]{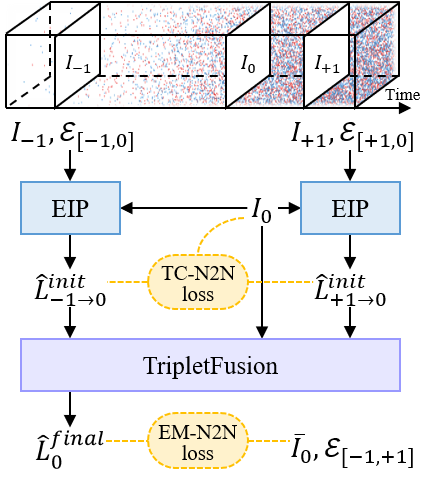} &
    \includegraphics[width=0.64\textwidth]{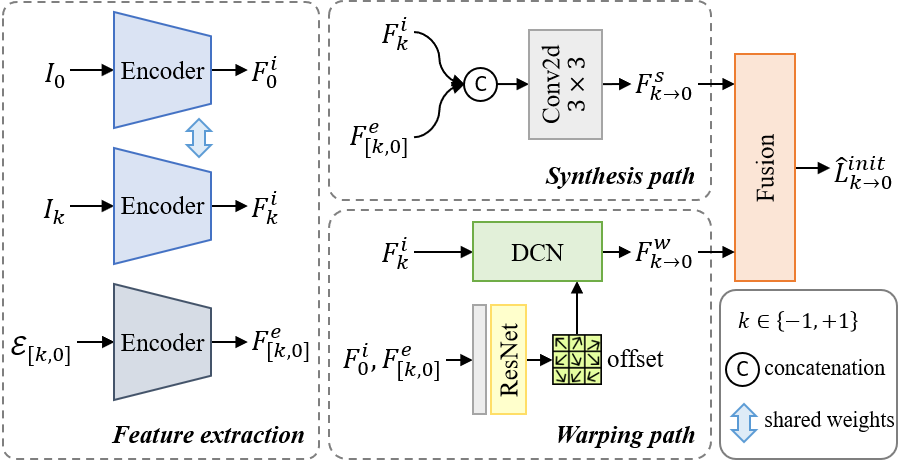} \\
    (a) Our \myback framework. & (b) Details of EIP module. \\
    \end{tabular}
    \caption{(a) Overview of our Self-supervised Event-based Image Denoising (\myback) framework. (b) Details of the Event-based Image Prediction (EIP) module.}
    \label{fig:seid}
\end{figure*}

As textures are highly correlated in the spatial domain, we define patch-wise counting to monitor the local motion information. Assuming that events have the spatial resolution of $H \times W$. Given the scale parameter $n$, we generate patches with the size $w \times w$, where $w=\frac{1}{n}\operatorname{min}(H,W)$, and calculate the patch-wise counting (\cref{fig:pea}(b)) as,
\begin{equation}
    \mathcal{C}(\mathbf{x},t;s,w) \doteq \sum_{e_i\in \Delta t}\delta(t-t_i)\delta(\mathbf{x}-\lfloor(\mathbf{x}_i+s)/w\rfloor),
\end{equation}
where $s$ is the padding size. Since the position $\mathbf{x}$ has been contained in each event data, the calculation of $\mathcal{C}(\cdot)$ still does not need any event-to-frame transformation, leading to an efficient EMM computation. 

To make patches overlap in implementation, we set $s\in\{0,w/2\}$, denoting by $\mathcal{C}(\mathbf{x},t;0,w)$ and $\mathcal{C}(\mathbf{x},t;w/2,w)$, separately. Thus, we define Local Event Accumulation (LEA) as,
\begin{equation}
    \begin{aligned}
        &\mathcal{M}_l(t;n)  \doteq \\ 
        &n \cdot \mathop{\operatorname{max}}\limits_{\mathbf{x}}\left(\int^t_0 \mathcal{C}(\mathbf{x},t;0,w)dt, \int^t_0 \mathcal{C}(\mathbf{x},t;w/2,w)dt\right).
    \end{aligned}
\end{equation}

As shown in \cref{fig:pea}(a), combining $\mathcal{M}_l(t;n)$ with multiple $n\in\{1,2,4\}$, which represent the multi-scale respective field, we propose the Pyramid Event Accumulation (PEA) to define EMM by,
\begin{equation}
    \mathcal{M}_\textit{p}(t) \doteq \mathop{\operatorname{max}}(\mathcal{M}_l(t;1), \mathcal{M}_l(t;2), \mathcal{M}_l(t;4)),
\end{equation}
where $\mathcal{M}_l(t;1)$ is actually the same as $\mathcal{M}_\textit{g}(t)$. Therefore, the proposed PEA can consider global and local blur simultaneously. As shown in \cref{fig:gea_vs_pea}, the instantaneous intensity frame received by the sensor is accompanied by destructive noise. Our \myfore\ with the GEA can help alleviate the noise by extending the exposure time in \cref{fig:gea_vs_pea}(b), while that with the PEA further achieves the less severe blurs and performs better by taking the local motion into account in \cref{fig:gea_vs_pea}(c).


\subsection{Self-supervised Event-based Image Denoising (\myback)}
\label{sec:seid}
The proposed \myfore\ primarily aims to effectively alleviate the motion blur in the frame acquisition process but does not address the degradation caused by the instant noise. In this section, we propose the Self-supervised Event-based Image Denoising (\myback) framework to further restore the latent noise-free image from the captured frame in a self-supervised learning manner.

\subsubsection{Image Denoising Network}
\label{sec:idn}
Considering the zero-mean characteristics of noise, we can filter it into a clean one by applying a simple integration. However it is prevented due to the blurry textures caused by the dynamic scenes. Fortunately, the inter-frame motion information recorded by event streams can help to compensate for the misalignment of neighboring frames. As shown in \cref{fig:seid}, our Image Denoising Network (IDN) contains two Event-based Image Prediction (EIP) modules and a TripletFusion module, fed with the target noisy image $I_0$, neighboring images $I_{-1},I_{+1}$, and inter-frame events $\mathcal{E}_{[-1,0]},\mathcal{E}_{[0,+1]}$. Note that before being fed to EIP, events between $I_0$ and $I_{+1}$ are reversed to $\mathcal{E}_{[+1,0]}$, similar to \cite{zhang2022unifying}.

EIP modules are designed to map the neighboring images $I_{-1/+1}$ to the target image $I_0$ with the assistance of the inter-frame events $\mathcal{E}_{[-1/+1,0]}$ as,
\begin{equation}
    \hat{L}^{init}_{k \rightarrow 0}=\operatorname{EIP}(I_0, I_k, \mathcal{E}_{[k,0]}),
    \label{eq:eip}
\end{equation}
where $k \in \{-1,+1\}$. Specifically, features of images and events are first extracted as $F^{i/e}=\operatorname{Encoder}(I/\mathcal{E})$. As shown in \cref{fig:seid}(b), inspired by the event-based frame interpolation methods \cite{tulyakov2021time}, we adopt dual paths, \ie, synthesis path and warping path, to alleviate non-alignment artifacts as,
\begin{equation}
    \begin{aligned}
        F^s_{k \rightarrow 0}&=\operatorname{SynthPath}(F^i_k,F^e_{k \rightarrow 0}), \\
        F^w_{k \rightarrow 0}&=\operatorname{WarpPath}(F^i_0,F^i_k,F^e_{k \rightarrow 0}), \\
    \end{aligned}
\end{equation}
where $F^i_0$ used in $\operatorname{WarpPath}(\cdot)$ is only for the calculation of the offset maps for a Deformable Convolution Network (DCN) block \cite{zhu2019deformable} and not directly for the $F^w_{k \rightarrow 0}$. The features $F^s_{k \rightarrow 0},F^w_{k \rightarrow 0}$ are then fused to predict the image $\hat{L}^{init}_{k \rightarrow 0}$. 

Finally, we utilize a UNet-based module as the TripletFusion module with the pre-aligned images $I_0, \hat{L}^{init}_{-1 \rightarrow 0}, \hat{L}^{init}_{+1 \rightarrow 0}$ as inputs to predict the final denoising result as,
\begin{equation}
    \hat{L}^{final}_0=\operatorname{TripletFusion}(I_0, \hat{L}^{init}_{-1 \rightarrow 0}, \hat{L}^{init}_{+1 \rightarrow 0}).
    \label{eq:l_final}
\end{equation}

\begin{figure*}[t]
    \centering
    \includegraphics[width=0.75\textwidth]{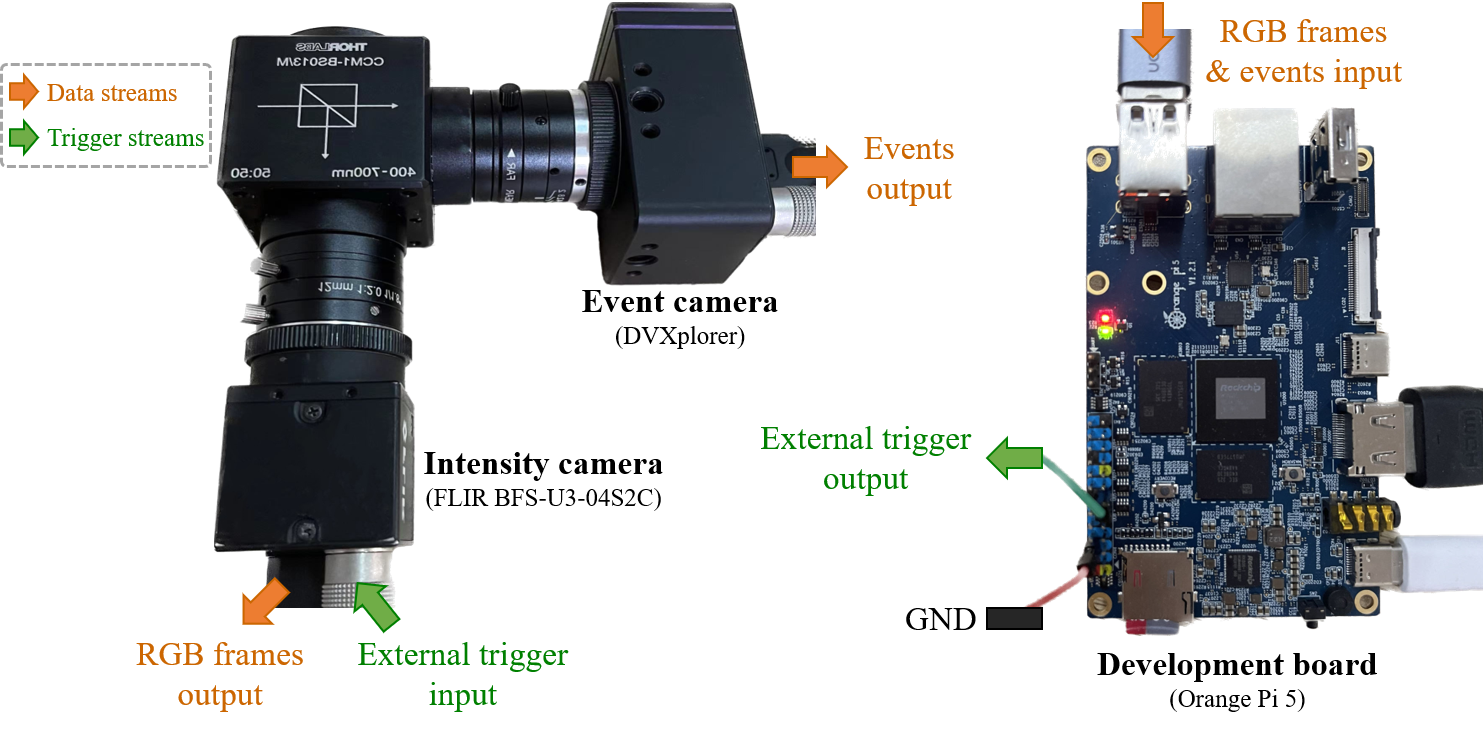}
    \caption{Our prototype system and the data streams and external trigger streams among cameras and the development board.}
    \label{fig:prototype}
\end{figure*}

\subsubsection{Self-supervised Learning}
We first revisit the loss function of existing image denoising methods, which can be formulated as,
\begin{equation}
    \mathcal{L}_{id}(I_{tar}) = \|\operatorname{IDN}(I_0)-I_{tar}\|_1,
\end{equation}
where $I_{tar}$ represents the target image, which is actually the ground-truth clean image $I_{gt}$ in the full-supervised learning framework. However, it is hard to collect the paired noisy and ground-truth images in real-world scenarios, especially in dynamic scenes, which motivates us to explore a self-supervised learning paradigm to train the denoising network.

With the zero-mean distribution assumption of noises, Lehtinen \etal, \cite{lehtinen2018noise2noise} propose Noise2Noise (N2N) loss $\mathcal{L}_{n2n}=\mathcal{L}_{id}(I'_0)$, where $I'_0$ is created by adding independent noise realization to the same clean image as $I_0$, for self-supervised training. Clearly, it is not feasible for a learned network to predict a different noisy image from another image. Thus, the network unavoidably converges to output the arithmetic mean of inputs for each pixel, \ie, $\mathbb{E}(\mathcal{L}_{n2n}(I'_0))=\mathbb{E}(\mathcal{L}_{id}(I_{gt}))+c$, where $c$ is a constant \cite{izadi2023image,vaksman2023patch}. However, obtaining such perfectly noisy pairs is challenging in real circumstances. On the other hand, several methods prove that adopting $I_{tar}=I_{0}$ with the Blind-Spot Network (BSN) \cite{krull2019noise2void} architecture as IDN to avoid identity mapping can achieve the unbiased estimation of the ground-truth noise-free image \cite{krull2019noise2void,laine2019high,huang2021neighbor2neighbor,lee2022ap}. However, these BSN-based methods suffer over-smooth distortions owing to the missing point-to-point mapping \cite{li2023spatially}.

\begin{algorithm}[t]
    \caption{\myfore$_\textit{g}$ implemented in our prototype.}
    \label{alg:protoytpe}
    \KwIn{Threshold $R$, maximum exposure time $T$.}
    \KwOut{Captured frames $\{I_k\}$, corresponding events $\{e_i\}$, exposure starting and ending times $\{t^s_k\}$, $\{t^e_k\}$.}
    
    Open event and intensity cameras, $k \leftarrow$ 1. \\
    Set Global Event Accumulation (GEA) $\mathcal{M}_\textit{g} \leftarrow$ 0. \\    
    \While{\textnormal{event camera is opened}}{
        External trigger level to intensity camera $\leftarrow$ 1; \\
        Exposure starting time $t^s_k\leftarrow$ current time $t$; \\
        \While{\textnormal{receive a slice containing $n$ events}}{
            $\mathcal{M}_\textit{g} += n$; \\
            \If{\textnormal{$\mathcal{M}_\textit{g} > R$ \textbf{or} $t-t^s_k > T$}}{
                External trigger level $\leftarrow$ 0; \\
                Save RGB frame $I_k$; $t^e_k\leftarrow$ current $t$; \\
                Reset GEA $\mathcal{M}_\textit{g} \leftarrow$ 0; $k++$; \\
                \textbf{break} \\
            }
        }
    }
    Normalize the pixel intensity of RGB frames $\{I_k\}$.
\end{algorithm}

Following N2N training paradigm, the proposed \myback framework utilizes a non-BSN architecture as IDN and generates artificial images as targets to form a self-supervised learning framework, containing two N2N loss functions, \ie, the Event-masked N2N Loss $\mathcal{L}_{em-n2n}$ and the Temporal Consistency N2N Loss $\mathcal{L}_{tc-n2n}$, as shown in \cref{fig:seid}(a).

\noindent\textbf{Event-masked N2N Loss.} We define the artificial target by $I_{tar}=\bar{I}_0$, where $\bar{I}_0$ is the exposure time-weighted average of the input frame $I_0$ and its two neighboring frames for less noises as,

\begin{equation}
    \bar{I}_0 = \frac{\sum^{+1}_{k=-1} T_k \cdot I_k}{\sum^{+1}_{k=-1} T_k}.
\end{equation}

Still, the unbiased estimation of the ground truth noise-free image remains, \ie, $\mathbb{E}(\mathcal{L}_{n2n}(\bar{I}_0))=\mathbb{E}(\mathcal{L}_{id}(I_{gt}))+c$, and the identity mapping is avoided. Note that the above $\mathcal{L}_{n2n}(\bar{I}_0)$ assumes the static scenes. To extend it to dynamic scenes, we harness the motion information inherently embedded in events to eliminate disturbances from dynamic scenes as,
\begin{equation}
    \mathcal{L}_{em-n2n} = \|\mathbb{M}(\mathcal{E}_{[-1,+1]})\cdot(\hat{L}^{final}_0-\bar{I}_0)\|_1,
    \label{eq:loss_em}
\end{equation}
where $\mathbb{M}(\cdot)$ represents a mask operator for the events emitted within the interval $[-1,+1]$, which assigns a value of 1 to static pixels (without emitted events) and a value of 0 to dynamic pixels (with emitted events).

\noindent\textbf{Temporal Consistency N2N Loss.} 
As mentioned in \cref{sec:idn}, given the event streams $\mathcal{E}_{[k,0]}$, the EIP module aims to spatially align and map the adjacent image $I_k$ to the target image $I_0$. The temporal consistency between the mapped image and the target image guides us to directly use the target image $I_0$ as the supervision signal for the EIP module. Besides, the independently distributed noises of $I_k$ and $I_0$ can also empower the EIP module the ability of denoising, leading to the Temporal Consistency N2N loss as,
\begin{equation}
    \mathcal{L}_{tc-n2n} = \frac{1}{2} \sum_{k\in\{-1,+1\}} \|\hat{L}^{init}_{k \rightarrow 0}-I_0\|_1.
    \label{eq:loss_tc}
\end{equation}

Finally, the total self-supervised learning framework can be summarized as,
\begin{equation}
    \mathcal{L} = \lambda_1\mathcal{L}_{em-n2n} + \lambda_2\mathcal{L}_{tc-n2n},
\end{equation}
with $\lambda_1$ and $\lambda_2$ denoting the balancing parameters.

\begin{figure*}[]
    \centering
    \begin{tabular}{c c}
        \hspace{-2mm}
        \includegraphics[width=0.66\textwidth]{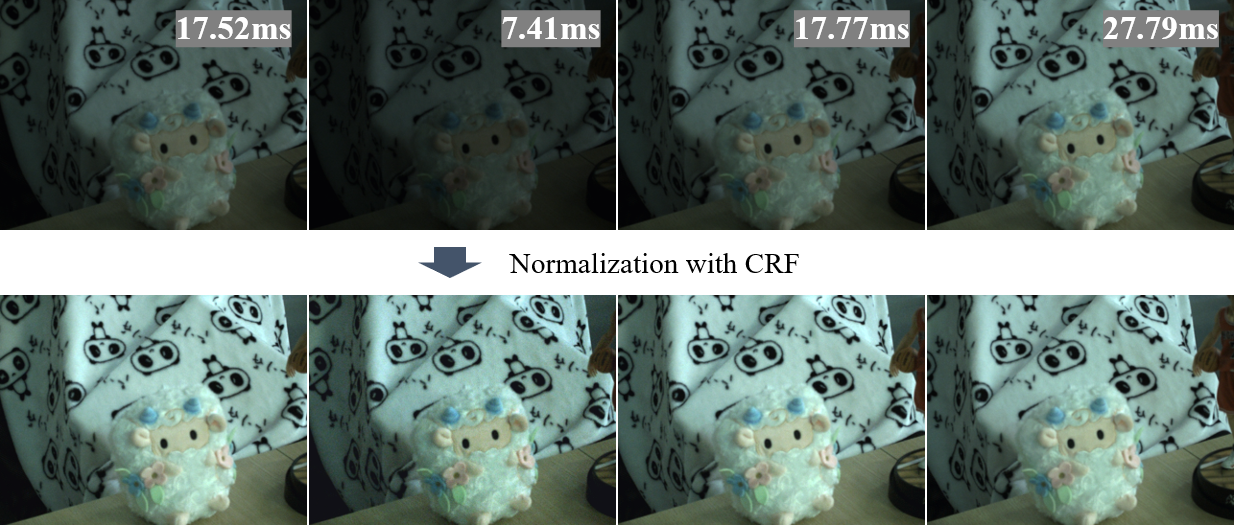} & \includegraphics[width=0.33\textwidth]{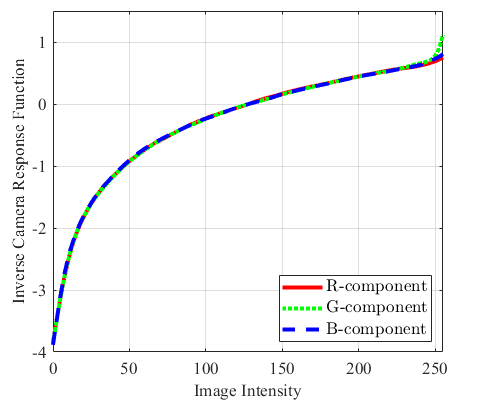} \\
        (a) Captured and normalized images by our prototype system. & (b) CRF calibration curves. \\
    \end{tabular}
    \caption{Intensity normalization results with the CRF calibration.}
    \label{fig:normalization}
\end{figure*}

\subsection{Prototype Building}
\label{sec:prototype}
\subsubsection{Hardware Composition}
To prove the efficiency of the proposed Neuromorphic Shutter Control (\myfore), we build a novel hybrid prototype system, as shown in \cref{fig:prototype}, to implement the proposed \myfore\ system using a conventional RGB camera (FLIR BFS-U3-04S2C), an event camera (DVXplorer), and a development board (Orange Pi 5). Specifically, the development board needs to calculate the EMM function, output the external trigger to the intensity camera, and receive the frame and event data from two cameras.

\subsubsection{Alignment} 
To ensure the same field of view of events and frames, we perform alignment using a homography estimated by matching SIFT features \cite{lowe2004distinctive}, which is computed using the RGB output from the FLIR camera and the gray-scale E2VID \cite{rebecq2019events} results of the event data. After alignment, the events and the RGB frames share the same spatial resolution of $640 \times 480$.

\subsubsection{Implementation of \myfore} 
The overall implementation process is summarized in \cref{alg:protoytpe}. We begin by initializing the threshold value $R$ and the maximum exposure time $T$ to avoid over-exposure. When the event camera is activated, the development board pulls up the level of the external trigger for the intensity camera to open the shutter of the intensity camera to start exposing. The development board receives the event slices and pulls down the level of the external trigger when the current Global Event Accumulation (GEA) $\mathcal{M}_\textit{g}$ reaches the threshold value $R$ or the current time period reaches the maximum exposure time $T$. Finally, the captured frame is saved for further processes. Although our Pyramid Event Accumulation (PEA) does not involve complicated calculations like spatial convolution, additional operations, such as event slice unpacking, make PEA unsuitable for embedding on the development board.

\subsubsection{Intensity Normalization}
In scenes with non-uniform motion, the frames captured by our prototype system have different exposure times, resulting in inconsistency in the intensity of the collected images in the temporal domain. With the relationship between exposure time $T$ and pixel intensity $I(T)$ developed by \cite{wang2022automated}, we can normalize the current intensity to the new intensity under the target exposure time $T_{tar}$ as,
\begin{equation}
    \mathcal{G}(I(T_{tar})) = \mathcal{G}(I(T)) + \ln(T_{tar}) - \ln(T),
\end{equation}
where $\mathcal{G}(\cdot)$ denotes the inverse Camera Response Function (CRF) calibrated by \cite{debevec2008recovering} and fitted with the tenth-order curve. The CRF calibration and intensity normalization results are shown in \cref{fig:normalization}. Note that the above equation ignores the influence of the other image signal processors, \eg, ``Auto Gain'' and ``Auto White Balance''. Thus, we disable such functions before capturing images. 

\def\ssxxsone{(-1.2,0.5)} 
\def\ssyysone{(-0.06,-2.05)} 
\def\ssxxstwo{(0.475,0.6)} 
\def\ssyystwo{(0.06,-2.05)} 
\def\ssmag{4}
\def\ssizz{1.65cm} 
\def\sswidth{0.19\textwidth} 
\begin{figure*}[t]
    \centering
    \begin{tabular}{c c c c c}
        \begin{tikzpicture}[spy using outlines={green,magnification=\ssmag,size=\ssizz},inner sep=0]
    		\node {\includegraphics[width=\sswidth]{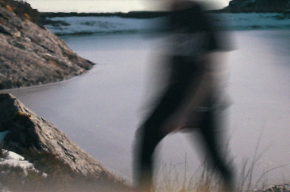}};
    		\spy on \ssxxsone in node [left] at \ssyysone;
            \spy [red] on \ssxxstwo in node [right,red] at \ssyystwo;
    	\end{tikzpicture} & \hspace{-4.5mm}
        \begin{tikzpicture}[spy using outlines={green,magnification=\ssmag,size=\ssizz},inner sep=0]
    		\node {\includegraphics[width=\sswidth]{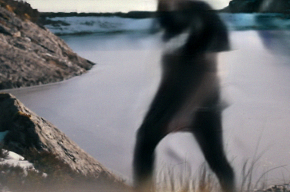}};
    		\spy on \ssxxsone in node [left] at \ssyysone;
            \spy [red] on \ssxxstwo in node [right,red] at \ssyystwo;
    	\end{tikzpicture} & \hspace{-4.5mm}
        \begin{tikzpicture}[spy using outlines={green,magnification=\ssmag,size=\ssizz},inner sep=0]
    		\node {\includegraphics[width=\sswidth]{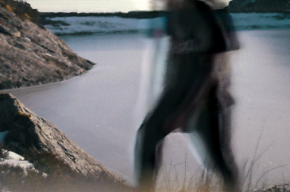}};
    		\spy on \ssxxsone in node [left] at \ssyysone;
            \spy [red] on \ssxxstwo in node [right,red] at \ssyystwo;
    	\end{tikzpicture} & \hspace{-4.5mm}
        \begin{tikzpicture}[spy using outlines={green,magnification=\ssmag,size=\ssizz},inner sep=0]
    		\node {\includegraphics[width=\sswidth]{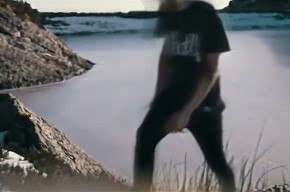}};
    		\spy on \ssxxsone in node [left] at \ssyysone;
            \spy [red] on \ssxxstwo in node [right,red] at \ssyystwo;
    	\end{tikzpicture} & \hspace{-4.5mm}
        \begin{tikzpicture}[spy using outlines={green,magnification=\ssmag,size=\ssizz},inner sep=0]
    		\node {\includegraphics[width=\sswidth]{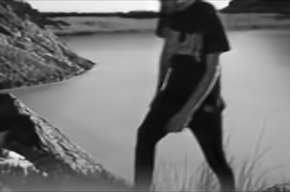}};
    		\spy on \ssxxsone in node [left] at \ssyysone;
            \spy [red] on \ssxxstwo in node [right,red] at \ssyystwo;
    	\end{tikzpicture} \\
        \hspace{-3mm}\small{(a) Full exposure} & \hspace{-3mm}\small{(b) METR \cite{zhang2021exposure}} & \hspace{-3mm}\small{(c) EFNet \cite{sun2022event}} & \hspace{-3mm}\small{(d) DCE \cite{zhang2023deep}} & \hspace{-3mm}\small{(e) DG \cite{dahary2021digital}} \\
        \begin{tikzpicture}[spy using outlines={green,magnification=\ssmag,size=\ssizz},inner sep=0]
    		\node {\includegraphics[width=\sswidth]{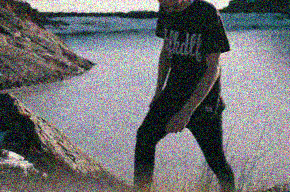}};
    		\spy on \ssxxsone in node [left] at \ssyysone;
            \spy [red] on \ssxxstwo in node [right,red] at \ssyystwo;
    	\end{tikzpicture} & \hspace{-4.5mm}
        \begin{tikzpicture}[spy using outlines={green,magnification=\ssmag,size=\ssizz},inner sep=0]
    		\node {\includegraphics[width=\sswidth]{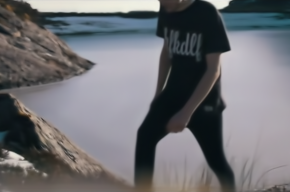}};
    		\spy on \ssxxsone in node [left] at \ssyysone;
            \spy [red] on \ssxxstwo in node [right,red] at \ssyystwo;
    	\end{tikzpicture} & \hspace{-4.5mm}
        \begin{tikzpicture}[spy using outlines={green,magnification=\ssmag,size=\ssizz},inner sep=0]
    		\node {\includegraphics[width=\sswidth]{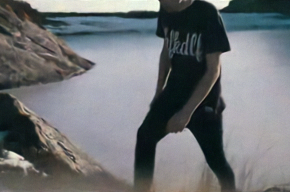}};
    		\spy on \ssxxsone in node [left] at \ssyysone;
            \spy [red] on \ssxxstwo in node [right,red] at \ssyystwo;
    	\end{tikzpicture} & \hspace{-4.5mm}
        \begin{tikzpicture}[spy using outlines={green,magnification=\ssmag,size=\ssizz},inner sep=0]
    		\node {\includegraphics[width=\sswidth]{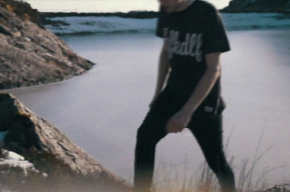}};
    		\spy on \ssxxsone in node [left] at \ssyysone;
            \spy [red] on \ssxxstwo in node [right,red] at \ssyystwo;
    	\end{tikzpicture} & \hspace{-4.5mm}
        \begin{tikzpicture}[spy using outlines={green,magnification=\ssmag,size=\ssizz},inner sep=0]
    		\node {\includegraphics[width=\sswidth]{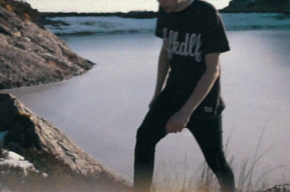}};
    		\spy on \ssxxsone in node [left] at \ssyysone;
            \spy [red] on \ssxxstwo in node [right,red] at \ssyystwo;
    	\end{tikzpicture} \\
        \hspace{-3mm}\small{(f) Instant frame} & \hspace{-3mm}\small{(g) PaCNet \cite{vaksman2021patch}} & \hspace{-3mm}\small{(h) PCST \cite{vaksman2023patch}} & \hspace{-3mm}\small{(i) \myfore$_\textit{g}$+\myback (ours)} & \hspace{-3mm}\small{(j) \myfore$_\textit{p}$+\myback (ours)} \\
    \end{tabular}
    \caption{Qualitative comparison on the Vimeo-Triplet dataset. Details are zoomed in for a better view.}
    \label{fig:result_vimeo}
\end{figure*}

\begin{table}
    \centering
    \caption{Quantitative comparisons of the proposed \myfore+\myback with various state-of-the-art methods on the Vimeo-Triplet dataset in terms of PSNR and SSIM. The best performance is in \textbf{bold} and the second best is \underline{underlined}.}
    \begin{tabular}{l l c c}
        \hline
        \multicolumn{2}{c}{\multirow{2}{*}{Method}} & \multicolumn{2}{c}{Vimeo-Triplet} \\
        \cline{3-4}
        \multicolumn{2}{c}{} & PSNR$\uparrow$ & SSIM$\uparrow$ \\
        \hline
        \multirow{2}{*}{Full exposure} & METR \cite{zhang2021exposure} & 23.022 & 0.707 \\
        & EFNet \cite{sun2022event} & 29.183 & 0.876 \\
        \hline
        \multirow{2}{*}{Uniform exposure} & PaCNet \cite{vaksman2021patch} & 34.995 & 0.957 \\
        & PCST \cite{vaksman2023patch} & 37.322 & 0.961 \\
        \hline
        \multirow{6}{*}{Non-uniform exposure} & DCE \cite{zhang2023deep} & 31.715 & 0.930 \\
        & DG \cite{dahary2021digital} & 34.771 & 0.943 \\
        & \myfore$_\textit{g}$+PCST \cite{vaksman2023patch} & 38.737 & 0.979 \\
        & \myfore$_\textit{p}$+PCST \cite{vaksman2023patch} & 38.742 & \underline{0.980} \\
        & \myfore$_\textit{g}$+\myback (ours) & \underline{39.247} & \textbf{0.983} \\
        & \myfore$_\textit{p}$+\myback (ours) & \textbf{39.260} & \textbf{0.983} \\
        \hline
    \end{tabular}
    \label{tab:main_comp}
\end{table}

\begin{figure}
    \centering
    \includegraphics[width=0.81\linewidth]{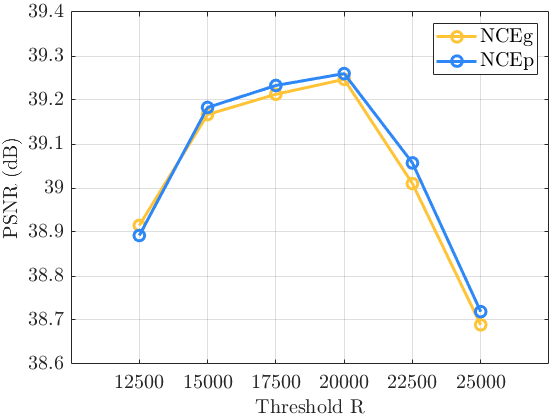}
    \caption{The curves of quantitative imaging results of our \myback\ with \myfore$_\textit{g}$ and \myfore$_\textit{p}$ under different thresholds $R$.}
    \label{fig:abla_diff_R}
\end{figure}

\section{Experiments}
In this section, we first evaluate the performance of our framework against state-of-the-art methods on one synthetic dataset in \cref{sec:exp_syn,sec:exp_abla}, and then capture a real-world dataset using the prototype system we built to validate the robustness and effectiveness of the proposed framework in real-world scenarios in \cref{sec:exp_real}. Finally, we analyze the time latency of the prototype system in \cref{sec:exp_time}.
\subsection{Comparisons on Synthetic Dataset}
\label{sec:exp_syn}
\subsubsection{Dataset} Since our framework produces and enhances non-uniform exposure images, it must be evaluated on dense radiance maps. Therefore, we follow \cite{zhang2022exploring} to construct our training and testing dataset based on the Vimeo-Triplet \cite{xue2019video}, containing a total of 2,500 sequences with 1,747 sequences for training and the other 753 sequences for testing. To avoid discontinuous artifacts, we first increase the frame rate of the original sequences with RIFE \cite{huang2022real}, finally owning 64 frames for each sequence, and generate events via DVS-Voltmeter \cite{lin2022dvs}. Then, we lower the pixel values for frames to simulate the instant intensity that the sensor achieves and add noises as \cref{eq:i_l+n}, where $\sigma_p$ and $\sigma_g$ are sampled from the uniform distribution $\mathcal{U}(0.01,0.04)$ as \cite{zhang2021learning}. 


\subsubsection{Implementation Details} Our non-uniform exposure imaging system contains an event-guided frame acquisition module \myfore\ and an image enhancement module \myback. We denote the proposed \myfore\ realized by GEA and PEA strategies by \myfore$_\textit{g}$ and \myfore$_\textit{p}$, respectively. For \myfore, the value of threshold $R$ is set to 20,000 by default, and we will discuss the influence of different $R$ on the final image results in \cref{sec:exp_abla}. For \myback, it is implemented using PyTorch and trained on a single NVIDIA Geforce RTX 3090 GPU with batch size 10 by default. During training, we randomly crop the samples into $256 \times 256$ patches. Adam optimizer is used for optimization, and the maximum epoch of training iterations is set to 60. The learning rate starts at $10^{-4}$, then decays by 50\% every 10 epochs from the 30-th epoch. The hyperparameters $\{\lambda_1, \lambda_2\}$ are set as: $\{1, 0.5\}$ by default.

\def\ssxxsone{(-0.1,0.5)} 
\def\ssyysone{(3.9,0)} 
\def\ssmag{3.5}
\def\ssizz{2.cm} 
\def\sswidth{0.2\textwidth} 
\begin{figure*}[]
    \centering
    \begin{tabular}{c c c}
        \vspace{-0.5mm}
        \hspace{-1mm}
        \begin{tikzpicture}[spy using outlines={green,magnification=\ssmag,size=\ssizz},inner sep=0]
    		\node {\includegraphics[width=\sswidth]{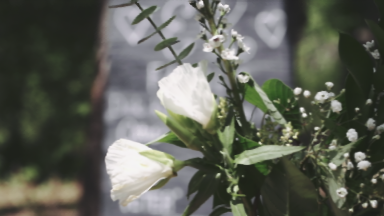}};
    		\spy on \ssxxsone in node [left] at \ssyysone;
    	\end{tikzpicture} \hspace{-4mm} & 
        \begin{tikzpicture}[spy using outlines={green,magnification=\ssmag,size=\ssizz},inner sep=0]
    		\node {\includegraphics[width=\sswidth]{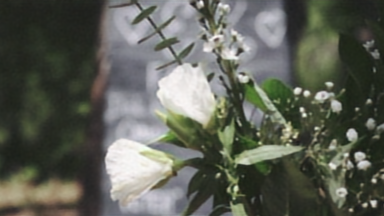}};
    		\spy on \ssxxsone in node [left] at \ssyysone;
    	\end{tikzpicture} \hspace{-4mm} & 
        \begin{tikzpicture}[spy using outlines={green,magnification=\ssmag,size=\ssizz},inner sep=0]
    		\node {\includegraphics[width=\sswidth]{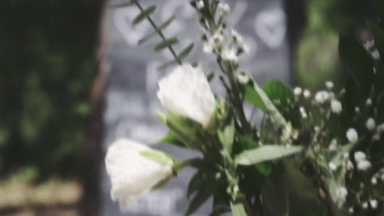}};
    		\spy on \ssxxsone in node [left] at \ssyysone;
    	\end{tikzpicture} \\
        \hspace{-1mm}
        \begin{tikzpicture}[spy using outlines={green,magnification=\ssmag,size=\ssizz},inner sep=0]
    		\node {\includegraphics[width=\sswidth]{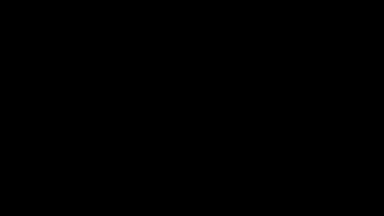}};
    		\spy on \ssxxsone in node [left] at \ssyysone;
    	\end{tikzpicture} \hspace{-4mm} & 
        \begin{tikzpicture}[spy using outlines={green,magnification=\ssmag,size=\ssizz},inner sep=0]
    		\node {\includegraphics[width=\sswidth]{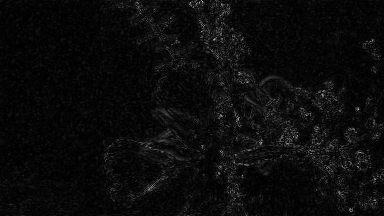}};
    		\spy on \ssxxsone in node [left] at \ssyysone;
    	\end{tikzpicture} \hspace{-4mm} & 
        \begin{tikzpicture}[spy using outlines={green,magnification=\ssmag,size=\ssizz},inner sep=0]
    		\node {\includegraphics[width=\sswidth]{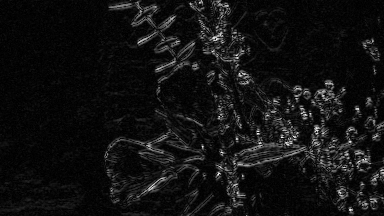}};
    		\spy on \ssxxsone in node [left] at \ssyysone;
    	\end{tikzpicture} \\
        \small{GT}\hspace{-4mm} & \small{w/ $\mathcal{L}_{tc-n2n}$, w/o $\mathcal{L}_{em-n2n}$}\textsuperscript{\textdagger}\hspace{-4mm} & \small{w/ $\mathcal{L}_{em-n2n}$ (w/ $\mathbb{M}$), w/o $\mathcal{L}_{tc-n2n}$} \vspace{-1mm} \\
        \small{(PSNR$\uparrow$, SSIM$\uparrow$)}\hspace{-4mm} & \small{(36.980, 0.986)}\hspace{-4mm} & \small{(31.937, 0.963)} \\
        \vspace{-0.5mm}
        \hspace{-1mm}
        \begin{tikzpicture}[spy using outlines={green,magnification=\ssmag,size=\ssizz},inner sep=0]
    		\node {\includegraphics[width=\sswidth]{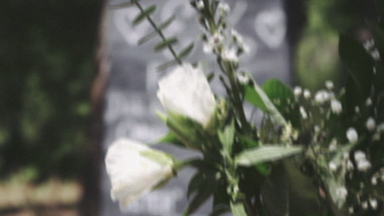}};
    		\spy on \ssxxsone in node [left] at \ssyysone;
    	\end{tikzpicture} \hspace{-4mm} & 
        \begin{tikzpicture}[spy using outlines={green,magnification=\ssmag,size=\ssizz},inner sep=0]
    		\node {\includegraphics[width=\sswidth]{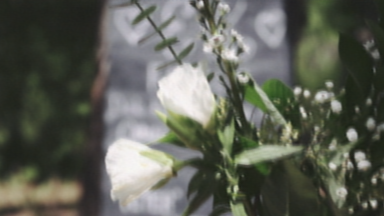}};
    		\spy on \ssxxsone in node [left] at \ssyysone;
    	\end{tikzpicture} \hspace{-4mm} & 
        \begin{tikzpicture}[spy using outlines={green,magnification=\ssmag,size=\ssizz},inner sep=0]
    		\node {\includegraphics[width=\sswidth]{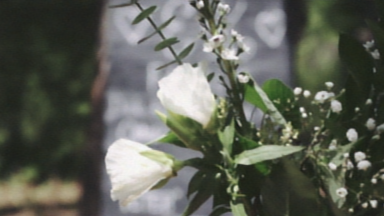}};
    		\spy on \ssxxsone in node [left] at \ssyysone;
    	\end{tikzpicture} \\
        \hspace{-1mm}
        \begin{tikzpicture}[spy using outlines={green,magnification=\ssmag,size=\ssizz},inner sep=0]
    		\node {\includegraphics[width=\sswidth]{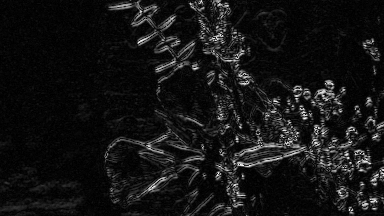}};
    		\spy on \ssxxsone in node [left] at \ssyysone;
    	\end{tikzpicture} \hspace{-4mm} & 
        \begin{tikzpicture}[spy using outlines={green,magnification=\ssmag,size=\ssizz},inner sep=0]
    		\node {\includegraphics[width=\sswidth]{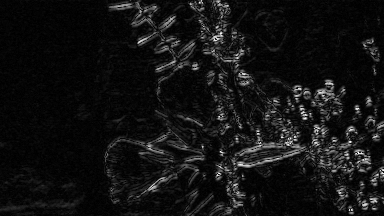}};
    		\spy on \ssxxsone in node [left] at \ssyysone;
    	\end{tikzpicture} \hspace{-4mm} & 
        \begin{tikzpicture}[spy using outlines={green,magnification=\ssmag,size=\ssizz},inner sep=0]
    		\node {\includegraphics[width=\sswidth]{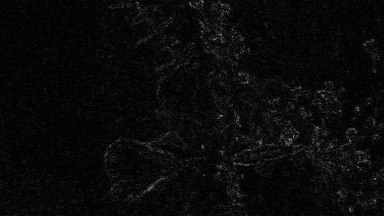}};
    		\spy on \ssxxsone in node [left] at \ssyysone;
    	\end{tikzpicture} \\
        \small{w/ $\mathcal{L}_{em-n2n}$ (w/o $\mathbb{M}$), w/o $\mathcal{L}_{tc-n2n}$}\hspace{-4mm} & \small{w/ $\mathcal{L}_{tc-n2n}+\mathcal{L}_{em-n2n}$ (w/o $\mathbb{M}$)}\hspace{-4mm} & \small{w/ all} \vspace{-1mm} \\
        \small{(30.115, 0.950)}\hspace{-4mm} & \small{(31.259, 0.962)}\hspace{-4mm} & \small{\textbf{(38.590, 0.988)}} \\
    \end{tabular}
    \caption{Qualitative ablations of each supervision and their absolute differences to the ground-truth clean image. \textdagger: We show the result of $\hat{L}^{init}_{+1 \rightarrow 0}$ instead of the final result for a better view.}
    \label{fig:abla_loss}
\end{figure*}

\begin{table}[]
    \centering
    \caption{Ablation study of supervisions on the Vimeo-Triplet dataset of the proposed Self-supervised Event-guided Image Denoising (SEID) framework. \textdagger: We evaluate the average matrices of the outputs of the EIP modules $\hat{L}^{init}_{-1 \rightarrow 0}$ and $\hat{L}^{init}_{+1 \rightarrow 0}$.}
    \begin{tabular}{c|c c c c c}
        \hline
        \multirow{2}{*}{Ex.} & \multirow{2}{*}{$\mathcal{L}_{tc-n2n}$} & \multicolumn{2}{c}{$\mathcal{L}_{em-n2n}$} & \multirow{2}{*}{PSNR$\uparrow$} & \multirow{2}{*}{SSIM$\uparrow$} \\
        \cline{3-4}
        & & w/o $\mathbb{M}(\cdot)$ & w/ $\mathbb{M}(\cdot)$ & & \\
        \hline
        1\textsuperscript{\textdagger} & \checkmark & & & 37.040 & 0.976 \\
        2 & & \checkmark & & 33.669 & 0.955 \\
        3 & & & \checkmark & 35.417 & 0.964 \\
        4 & \checkmark & \checkmark & & 34.150 & 0.960 \\
        5 & \checkmark & & \checkmark & \textbf{39.260} & \textbf{0.983} \\
        \hline
    \end{tabular}
    \label{tab:abla}
\end{table}


\subsubsection{Qualitative and Quantitative Comparisons}
We compare our system with various methods based on three different imaging strategies, \ie, motion deblurring methods applied on a full-exposure image (METR \cite{zhang2021exposure} and EFNet \cite{sun2022event}), image denoising methods on uniform-exposure images (PaCNet \cite{vaksman2021patch} and PCST \cite{vaksman2023patch}), and non-uniform exposure and enhancement methods (DG \cite{dahary2021digital} and DCE \cite{zhang2023deep}), in terms of Peak Signal to Noise Ratio (PSNR, higher is better) and Structural SIMilarity (SSIM, higher is better) \cite{wang2004image}. For a fair comparison, all the methods share the same overall budget of sequences.

Qualitative and quantitative results on the Vimeo-Triplet dataset are presented in \cref{fig:result_vimeo,tab:main_comp}. Overall, the proposed neuromorphic exposure imaging system \myfore+\myback outperforms the state-of-the-art methods by a large margin in terms of PSNR and SSIM. In the motion deblurring task, both the framed-based method METR \cite{zhang2021exposure} and event-based method EFNet \cite{sun2022event} utilize the full-exposure image as the input. Due to the motion-induced color bleeding and aliasing effects, they fail to reconstruct sharp and reliable textures. A clear example is the T-shirt within the red box in \cref{fig:result_vimeo}. Conversely, reducing the exposure time can prevent blur but results in a noisy instantaneous frame. Since the image denoising methods, \ie, PaCNet \cite{vaksman2021patch} and PCST \cite{vaksman2023patch}, are fed with the instant frame that does not record motion implicitly in the frame acquisition process, they achieve relatively higher PSNR and SSIM than the motion deblurring methods, as demonstrated in \cref{tab:main_comp}. However, the image denoising methods tend to over-smooth results and erase the original texture information, \eg, the hillside within the green box of \cref{fig:result_vimeo}.

For the non-uniform exposure imaging task, DCE \cite{zhang2023deep} is a coded exposure deblurring method with a one-shot image with the fluttering shutter as the input. Hence, the performance of DCE is limited by the lost intra-frame information. As a burst imaging method with learnable exposure times, DG \cite{dahary2021digital} is affected by artifacts introduced in the multi-frame alignment. As shown in \cref{fig:result_vimeo} (i) and (j), our neuromorphic exposure imaging system, with the \myfore\ to monitor the real-time motion information and the \myback to estimate latent noise-free images, can achieve clean and sharp results, avoiding motion blur while retaining original textures. Specifically, \myfore$_\textit{p}$ can handle the local motion in the frame acquisition process, resulting in sharper edges than \myfore$_\textit{g}$, \eg, \cref{fig:result_vimeo} with the static camera and the moving person. 

To further prove the effectiveness of our \myfore\ and \myback, we combine the proposed \myfore\ (GEA and PEA strategies) with the state-of-the-art image denoising method PCST \cite{vaksman2023patch}, \ie, \myfore$_\textit{g}$+PCST and \myfore$_\textit{p}$+PCST. Quantitative results in \cref{tab:main_comp} draw two conclusions: 1) The front-end \myfore\ imaging strategy of our algorithm is capable of producing images with a higher SNR compared to instant noisy images, reducing the burden of the back-end image denoising algorithm; 2) With the same inputs provided by \myfore, our \myback still outperforms PCST \cite{vaksman2023patch},  quantitatively validating our algorithm.

\def\ssxxsone{(0.8,0.5)} 
\def\ssyysone{(3.1,0)} 
\def\ssxxstwo{(-0.1,-0.1)} 
\def\ssyystwo{(3.1,0)} 
\def\ssxxsthree{(0.5,-0.1)} 
\def\ssyysthree{(3.1,0)} 
\def\ssmag{4}
\def\ssizz{1.8cm} 
\def\sswidth{0.275\linewidth} 
\begin{figure}
    \centering
    \begin{tabular}{c c}
        \hspace{-2mm}
        \begin{tikzpicture}[spy using outlines={red,magnification=\ssmag,size=\ssizz},inner sep=0]
    		\node {\includegraphics[width=\sswidth]{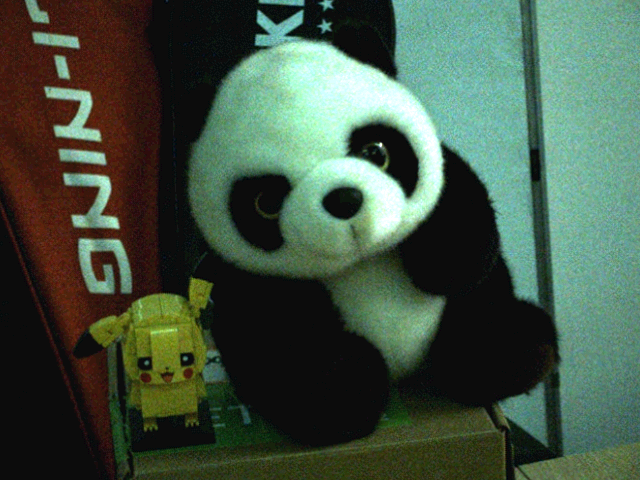}};
    		\spy on \ssxxsone in node [left] at \ssyysone;
    	\end{tikzpicture} & \hspace{-4.5mm}
        \begin{tikzpicture}[spy using outlines={green,magnification=\ssmag,size=\ssizz},inner sep=0]
    		\node {\includegraphics[width=\sswidth]{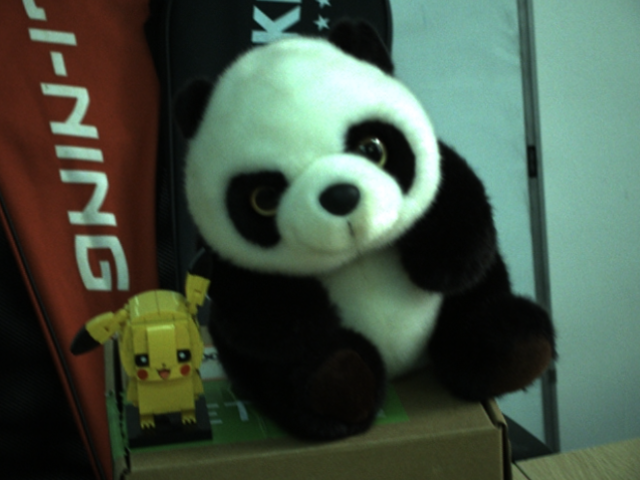}};
    		\spy on \ssxxsone in node [left] at \ssyysone;
    	\end{tikzpicture} \vspace{-1.5mm} \\
        \hspace{-3mm}\footnotesize{Noisy frame with short exposure} & \hspace{-3mm}\footnotesize{Frame captured with our \myfore} \vspace{-1.mm} \\
        \multicolumn{2}{c}{\small (a) \textit{Static scene} (with static camera and object)} \\
        \hspace{-2mm}
        \begin{tikzpicture}[spy using outlines={red,magnification=\ssmag,size=\ssizz},inner sep=0]
    		\node {\includegraphics[width=\sswidth]{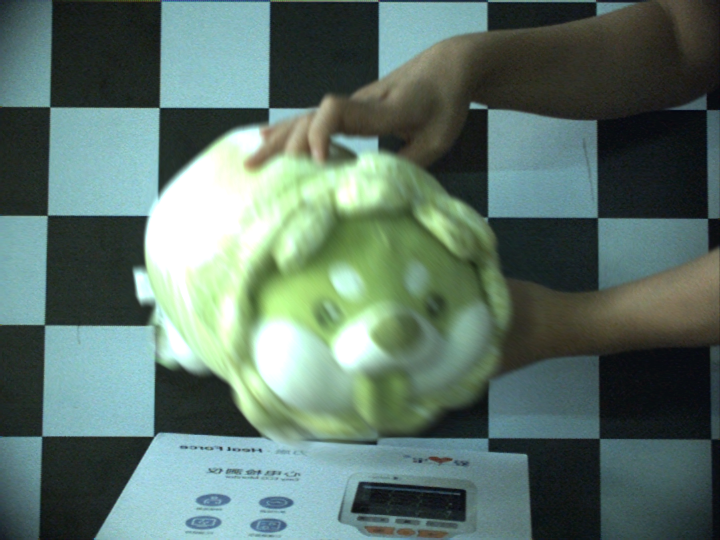}};
    		\spy on \ssxxstwo in node [left] at \ssyystwo;
    	\end{tikzpicture} & \hspace{-4.5mm}
        \begin{tikzpicture}[spy using outlines={green,magnification=\ssmag,size=\ssizz},inner sep=0]
    		\node {\includegraphics[width=\sswidth]{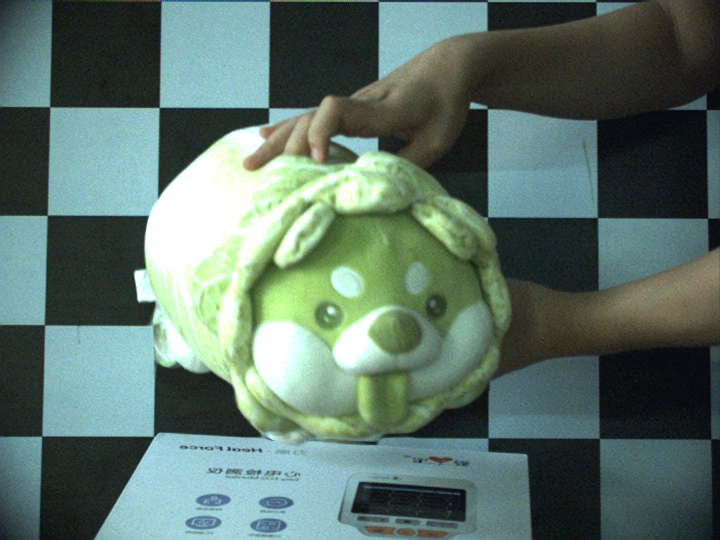}};
    		\spy on \ssxxstwo in node [left] at \ssyystwo;
    	\end{tikzpicture} \vspace{-1.5mm} \\
        \hspace{-3mm}\footnotesize{Blurry frame with long exposure} & \hspace{-3mm}\footnotesize{Frame captured with our \myfore} \vspace{-1.mm} \\
        \multicolumn{2}{c}{\small (b) \textit{Dynamic object} (with static camera)} \\
        \hspace{-2mm}
        \begin{tikzpicture}[spy using outlines={red,magnification=\ssmag,size=\ssizz},inner sep=0]
    		\node {\includegraphics[width=\sswidth]{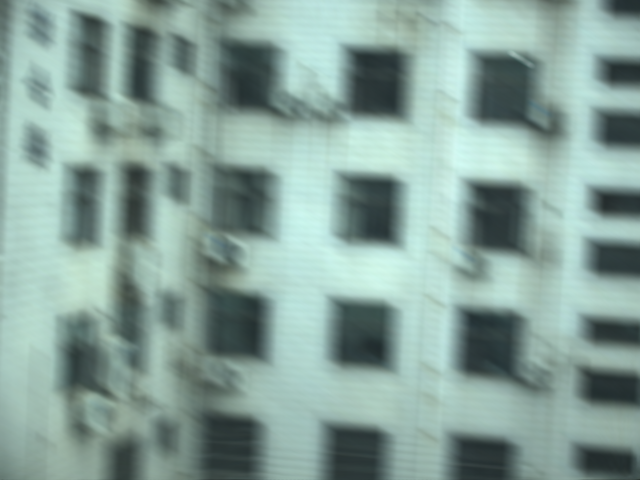}};
    		\spy on \ssxxsthree in node [left] at \ssyysthree;
    	\end{tikzpicture} & \hspace{-4.5mm}
        \begin{tikzpicture}[spy using outlines={green,magnification=\ssmag,size=\ssizz},inner sep=0]
    		\node {\includegraphics[width=\sswidth]{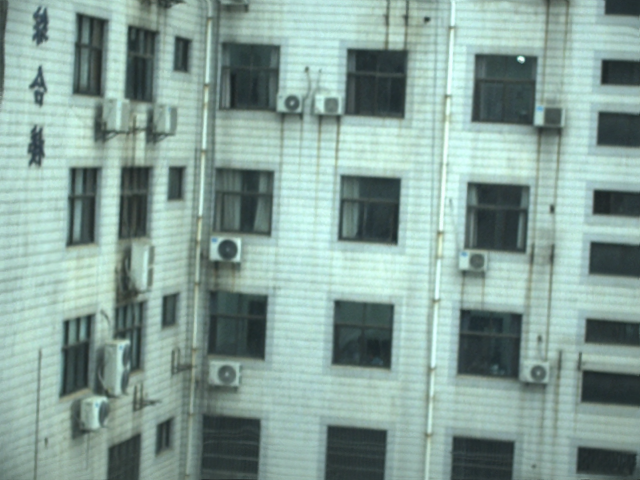}};
    		\spy on \ssxxsthree in node [left] at \ssyysthree;
    	\end{tikzpicture} \vspace{-1.5mm} \\
        \hspace{-3mm}\footnotesize{Blurry frame with long exposure} & \hspace{-3mm}\footnotesize{Frame captured with our \myfore} \vspace{-1.mm} \\
        \multicolumn{2}{c}{\small (c) \textit{Dynamic camera} (with static or dynamic object)}
    \end{tabular}
    \caption{Examples on our dataset \mydata\ in different motion patterns, proving the efficiency of our \myfore\ in avoiding motion blurs and noises. All images are normalized for better viewing.}
    \label{fig:dataset}
\end{figure}

\subsection{Ablation Studies}
\label{sec:exp_abla}
\subsubsection{Influence of the Threshold R}
Firstly, we discuss the influence of different values of the threshold $R$ on the imaging results of the proposed \myfore$_\textit{g}$ and \myfore$_\textit{p}$. A higher threshold signifies a greater impact of motion blur on the captured image. In comparison, a lower threshold yields a sharper image but may introduce another form of corruption, namely noise. As shown in \cref{fig:abla_diff_R}, \myfore+\myback with $R=20,000$ seeks out the balance of the noise-blur trade-off on the Vimeo-Triplet \cite{xue2019video} dataset, achieving the highest PSNR score. Moreover, with the same threshold $R$, \myfore$_\textit{p}$+\myback basically achieves better performance than \myfore$_\textit{g}$+\myback does by taking the local motion into account, except when $R=12,500$ due to the dominant noise caused by the extremely short exposure time.

\begin{figure}[]
    \centering
    \includegraphics[width=\linewidth]{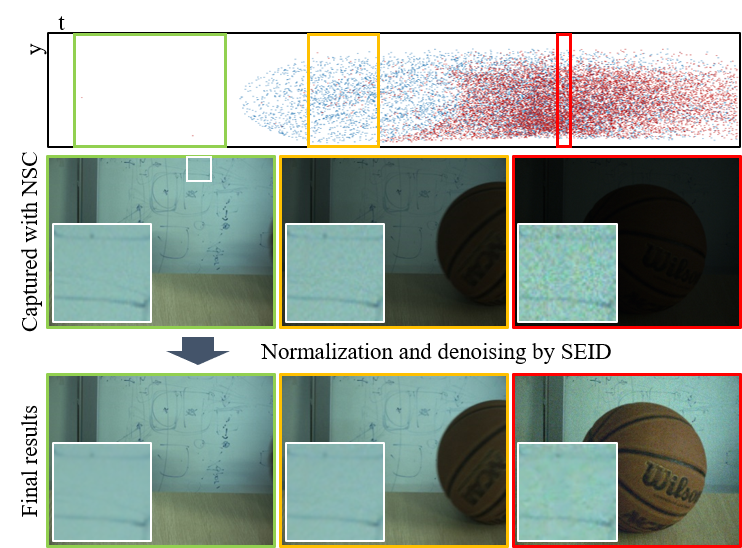}
    \caption{Three sample frames of the sequence ``basketball\underline{~~}8'' containing a rolling basketball on our dataset \mydata collected with the proposed \myfore\ and the imaging results by \myback.}
    \label{fig:sample}
\end{figure}

\begin{table}[]
    \centering
    \caption{Overview of our \mydata. \#Frame is the total number of RGB frames and \#Event (M) indicates the total number of events of each motion pattern.}
    \begin{tabularx}{\linewidth}{l|p{0.68cm}<{\centering} p{1.39cm}<{\centering} X<{\raggedright\arraybackslash}}
        \hline
        Motion patterns & \#Frame & \#Event (M) & Sequences\\
        \hline
        \textit{Static scene} & 500 & 0.03 & static\underline{~~}1$\sim$6, building\underline{~~}1$\sim$4 \\
        \hline
        \multirow{3}{*}{\textit{Dynamic object}} & \multirow{3}{*}{1,000} & \multirow{3}{*}{31.23} & basketball\underline{~~}3$\sim$8, dog\underline{~~}2, checkerboard\underline{~~}1$\sim$2, walker\underline{~~}1$\sim$2 \\
        \hline
        \multirow{5}{*}{\textit{Dynamic camera}} & \multirow{5}{*}{1,950} & \multirow{5}{*}{125.81} & panda\underline{~~}1$\sim$4, jacket\underline{~~}1$\sim$2, table\underline{~~}1$\sim$6, blanket\underline{~~}1$\sim$3, building\underline{~~}5$\sim$9, bucket\underline{~~}1, basketball\underline{~~}1$\sim$2, case\underline{~~}1, scenery\underline{~~}1$\sim$5, dog\underline{~~}1 \\
        \hline
    \end{tabularx}
    \label{tab:data_detail}
\end{table}

\subsubsection{Necessity of Loss Combination}
From the absolute difference to the ground-truth image and the metric performance (\ie, PSNR$\uparrow$/SSIM$\uparrow$) shown in \cref{fig:abla_loss}, we can observe that removing either $\mathcal{L}_{tc-n2n}$ or $\mathcal{L}_{em-n2n}$ leads to a degradation of the prediction of the results, indicating their contribution in generating the clean and sharp image. Specifically, $\mathcal{L}_{tc-n2n}$ directly introduces the determined input, \ie, the target noisy frame as the strong supervision to guide the EIP modules to learn spatial and temporal transitions. On the foundation of the constraint of brightness consistency, $\mathcal{L}_{em-n2n}$ guides \myback to predict the final denoised results, where $\mathbb{M}(\cdot)$ utilizes the event data to help filter out many moving textures that cause blur and retain stationary textures as supervision signals. In general, \cref{tab:abla} and \cref{fig:abla_loss} show that combining all these losses leads to the smallest absolute error, validating the necessity of recovering latent clean images with $\mathcal{L}_{tc-n2n}$ and $\mathcal{L}_{em-n2n}$ simultaneously. Note that since only applying $\mathcal{L}_{tc-n2n}$ prevents the TripletFusion module from converging, we evaluate the average matrices of the outputs of the EIP modules $\hat{L}^{init}_{-1 \rightarrow 0}$ and $\hat{L}^{init}_{+1 \rightarrow 0}$ for Ex.1 in \cref{tab:abla}.

\def\ssxxsone{(0.4,0.75)} 
\def\ssyysone{(4.25,0)} 
\def\ssxxstwo{(1.28,-0.6)} 
\def\ssyystwo{(4.25,0)} 
\def\ssmag{3.5}
\def\ssizz{2.52cm} 
\def\sswidth{0.185\textwidth} 
\begin{figure*}[t]
    \centering
    \begin{tabular}{c c c}
        \hspace{-3mm}
        \begin{tikzpicture}[spy using outlines={green,magnification=\ssmag,size=\ssizz},inner sep=0]
    		\node {\includegraphics[width=\sswidth]{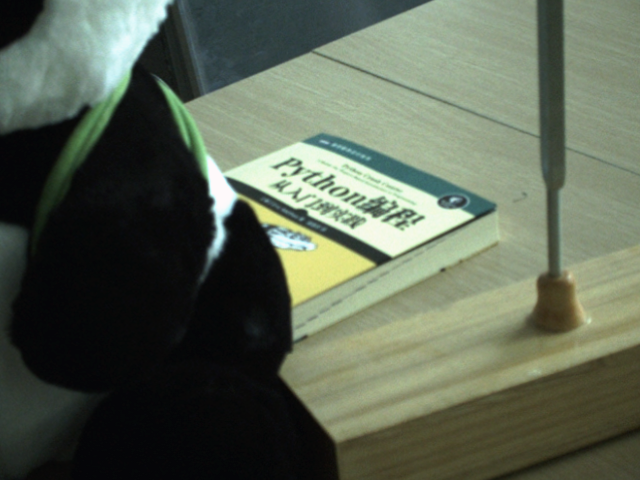}};
    		\spy on \ssxxsone in node [left] at \ssyysone;
    	\end{tikzpicture} & \hspace{-4mm}
        \begin{tikzpicture}[spy using outlines={green,magnification=\ssmag,size=\ssizz},inner sep=0]
    		\node {\includegraphics[width=\sswidth]{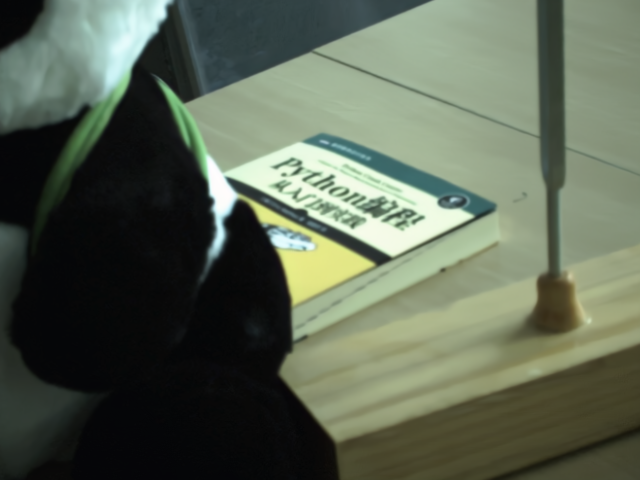}};
    		\spy on \ssxxsone in node [left] at \ssyysone;
    	\end{tikzpicture} & \hspace{-4mm}
        \begin{tikzpicture}[spy using outlines={green,magnification=\ssmag,size=\ssizz},inner sep=0]
    		\node {\includegraphics[width=\sswidth]{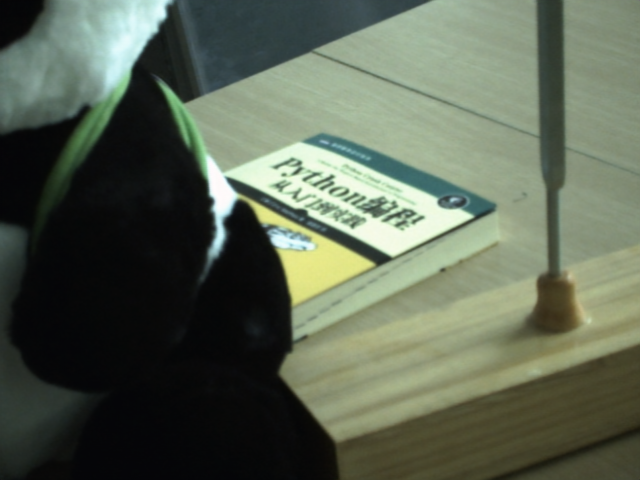}};
    		\spy on \ssxxsone in node [left] at \ssyysone;
    	\end{tikzpicture} \\
    \hspace{-3mm}
        \begin{tikzpicture}[spy using outlines={green,magnification=\ssmag,size=\ssizz},inner sep=0]
    		\node {\includegraphics[width=\sswidth]{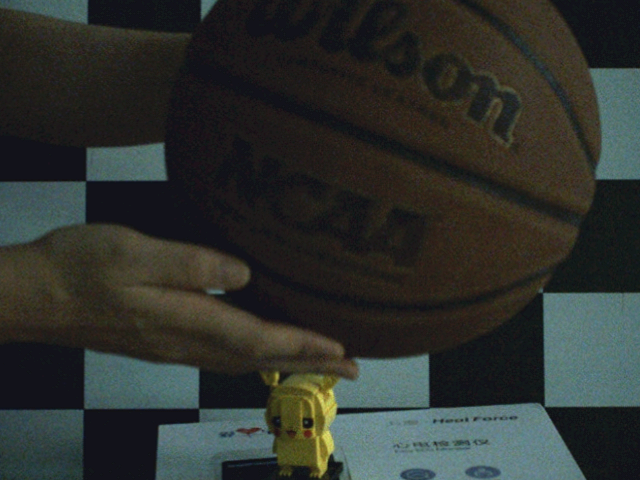}};
    		\spy on \ssxxstwo in node [left] at \ssyystwo;
    	\end{tikzpicture} & \hspace{-4mm}
        \begin{tikzpicture}[spy using outlines={green,magnification=\ssmag,size=\ssizz},inner sep=0]
    		\node {\includegraphics[width=\sswidth]{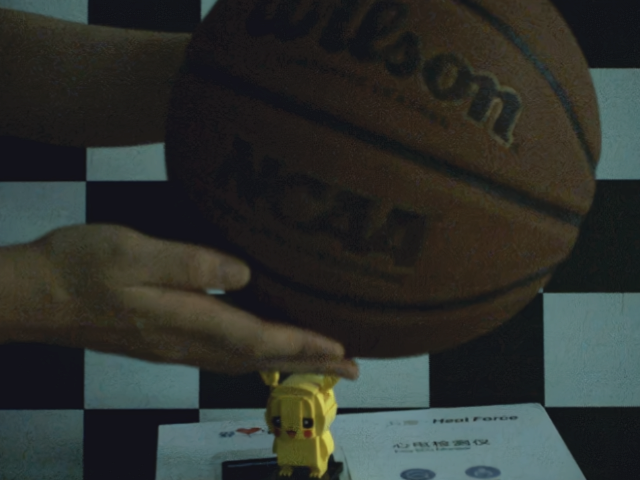}};
    		\spy on \ssxxstwo in node [left] at \ssyystwo;
    	\end{tikzpicture} & \hspace{-4mm}
        \begin{tikzpicture}[spy using outlines={green,magnification=\ssmag,size=\ssizz},inner sep=0]
    		\node {\includegraphics[width=\sswidth]{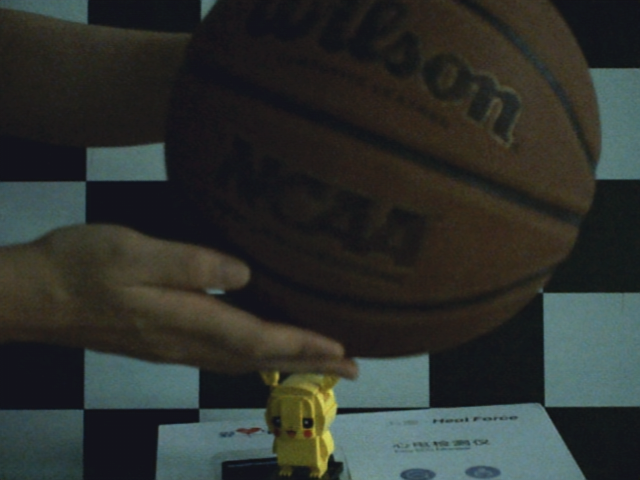}};
    		\spy on \ssxxstwo in node [left] at \ssyystwo;
    	\end{tikzpicture} \\
        \hspace{-3mm}(a) Captured frame by our prototype. & \hspace{-3mm}(b) PCST \cite{vaksman2023patch} & \hspace{-3mm}(c) \myback (ours) \\
    \end{tabular}
    \vspace{-2mm}
    \caption{Qualitative comparisons of the proposed \myback with a state-of-the-art image denoising method PCST \cite{vaksman2023patch} in two sequences ``panda\underline{~~}3'' (upper row) and ``basketball\underline{~~}3'' (bottom row) from our dataset \mydata. }
    \vspace{-2mm}
    \label{fig:result_ned}
\end{figure*}

\def\ssyys{(-1.5,0.9)}
\begin{figure*}[]
    \centering
    \begin{tabular}{c c}
        \includegraphics[width=0.46\textwidth]{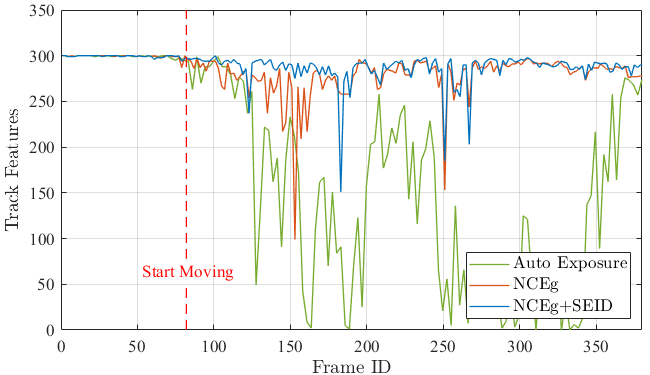} \hspace{-3mm} &
        \begin{minipage}[b]{0.565\textwidth}
            \begin{tikzpicture}[inner sep=0]
                \node {\includegraphics[width=0.3\textwidth]{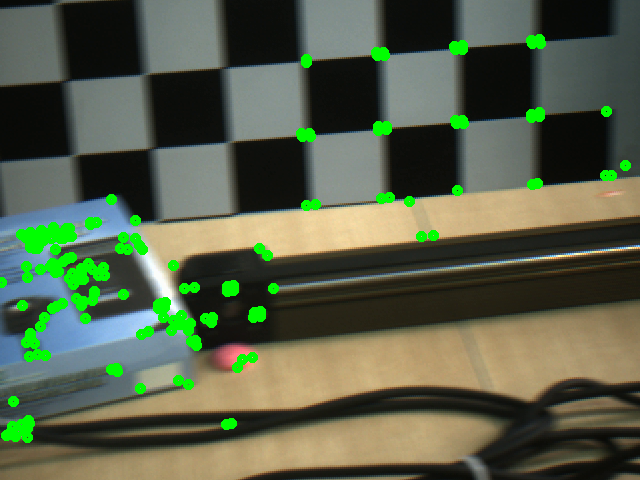}};
                \node [anchor=west] at (-1.5,0.975) {\textcolor{orange}{\bf AE}};
            \end{tikzpicture} \hspace{-2mm}
            \begin{tikzpicture}[inner sep=0]
                \node {\includegraphics[width=0.3\textwidth]{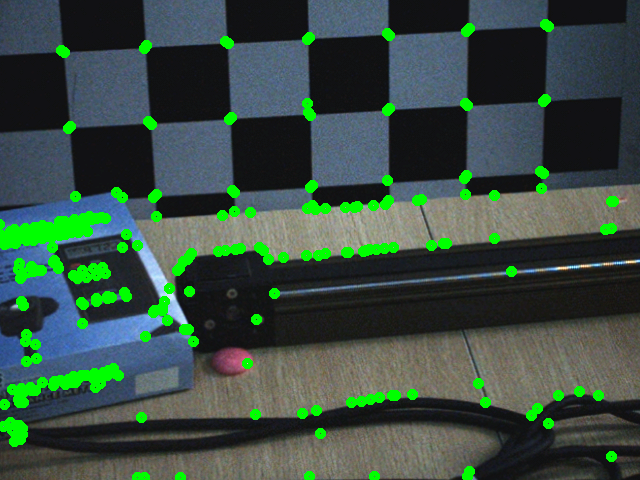}};
                \node [anchor=west] at \ssyys {\textcolor{orange}{\bf \myfore$_\textit{g}$}};
            \end{tikzpicture} \hspace{-2mm}
            \begin{tikzpicture}[inner sep=0]
                \node {\includegraphics[width=0.3\textwidth]{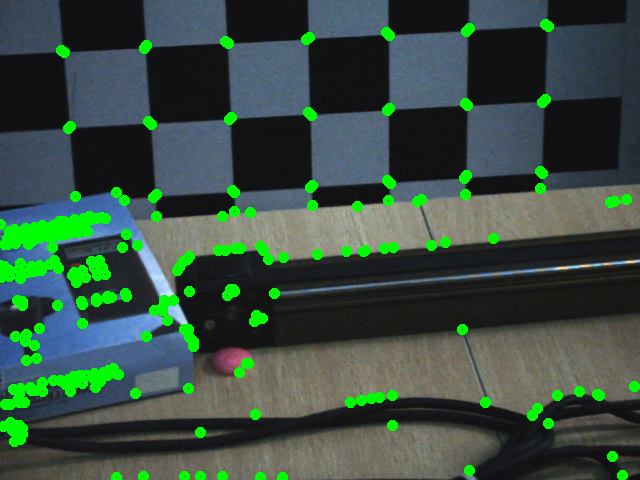}};
                \node [anchor=west] at \ssyys {\textcolor{orange}{\bf \myfore$_\textit{g}$+\myback}};
            \end{tikzpicture} \vspace{1mm} \\
            \begin{tikzpicture}[inner sep=0]
                \node {\includegraphics[width=0.3\textwidth]{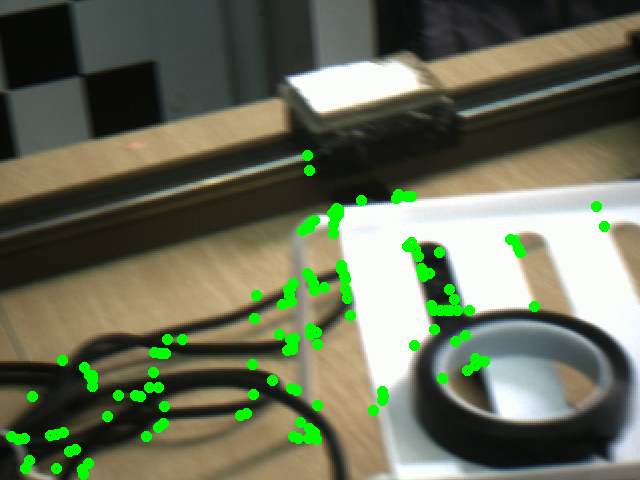}};
                \node [anchor=west] at (-1.5,0.975) {\textcolor{orange}{\bf AE}};
            \end{tikzpicture} \hspace{-2mm}
            \begin{tikzpicture}[inner sep=0]
                \node {\includegraphics[width=0.3\textwidth]{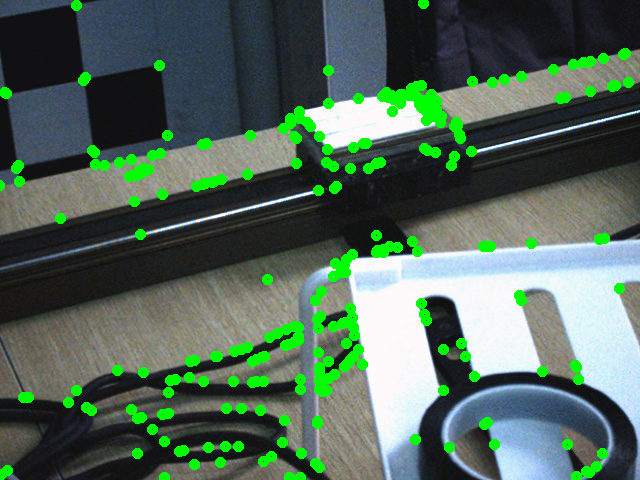}};
                \node [anchor=west] at \ssyys {\textcolor{orange}{\bf \myfore$_\textit{g}$}};
            \end{tikzpicture} \hspace{-2mm}
            \begin{tikzpicture}[inner sep=0]
                \node {\includegraphics[width=0.3\textwidth]{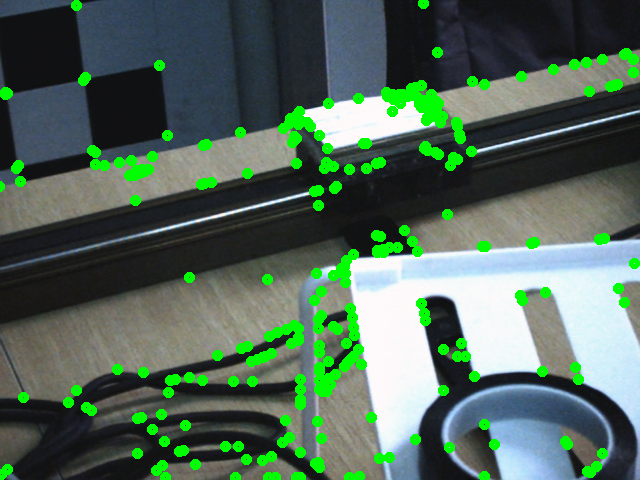}};
                \node [anchor=west] at \ssyys {\textcolor{orange}{\bf \myfore$_\textit{g}$+\myback}};
            \end{tikzpicture}
        \end{minipage} \\
        (a) Trend of the number of tracking feature points. & \hspace{-5mm}(b) Qualitative comparisons of the tracked features. \\
    \end{tabular}
    \caption{KLT\cite{lucas1981iterative,tomasi1991detection} feature tracking results with various imaging strategies. It could be clearly observed that tracking using blurry images captured with Auto Exposure (AE) cannot extract effective features. In contrast, tracking using the proposed \myfore$_\textit{g}$ and \myback has much more effective features and is more stable.}
    \label{fig:klt}
\end{figure*}

\begin{table}[]
    \centering
    \caption{Statistical analysis of KLT feature tracking.}
    \begin{tabular}{l c c c}
        \hline
        \multirow{2}{*}{Method} & \multicolumn{3}{c}{Tracking Feature Points} \\
        \cline{2-4}
        & Min$\uparrow$ & Mean$\uparrow$ & Median$\uparrow$ \\
        \hline
        Auto Exposure & 0 & 192 & 209 \\
        \myfore$_\textit{g}$ (ours) & 99 & 283 & 286 \\
        \myfore$_\textit{g}$+\myback (ours) & \textbf{151} & \textbf{290} & \textbf{292} \\
        \hline
    \end{tabular}
    \label{tab:klt}
\end{table}

\subsection{Demonstration in Real-world Scenarios}
\label{sec:exp_real}
\subsubsection{Dataset Collection}
With the prototype system we built in \cref{sec:prototype}, we collect a real-world dataset, named Neuromorphic Exposure Dataset (\mydata), containing the frames captured with the \myfore\ and corresponding events. We consider two classes of motions that are common in real life: object motion and camera motion. Our \mydata includes 51 sequences in three types of combination of object and camera motion, \ie, \textit{static scene} (with static camera and object), \textit{dynamic object} (with static camera), and \textit{dynamic camera} (with static or dynamic object), as shown in \cref{fig:dataset,tab:data_detail}. With the assistance of the event camera, our proposed \myfore\ can exploit the optimal noise-blur trade-off by actively extending exposure time in static scenes to avoid noises (\cref{fig:dataset}(a)) and shortening the exposure time in dynamic scenes to avoid blurs (\cref{fig:dataset}(b) and (c)). \mydata encompasses a wide range of real-world scenes, including simple indoor objects as well as complex outdoor construction sites.

\subsubsection{Validation in Dynamic Scenes} 
\cref{fig:sample} depicts a sample sequence that records the transformation of a static scene to a dynamic one due to an invading basketball, leading to a sudden surge in the number of events. In response, \myfore\ proactively shortens the exposure time to maintain sharp imaging. Additionally, our \myback effectively eliminates residual noises and stabilizes the Signal-to-Noise Ratio (SNR) of the captured frames, thereby producing high-quality results. Besides, we conduct a qualitative comparison between our image denoising algorithm, \myback, and the recent state-of-the-art method PCST \cite{vaksman2023patch} using our real-world dataset, \mydata. As illustrated in the bottom row of \cref{fig:result_ned}, both \myback and PCST exhibit similar capabilities in denoising non-textured regions of images. However, PCST inevitably erases the original textures present in the images, \eg, the table in the upper row of \cref{fig:result_ned}. This unexpected outcome leads to a reduction in the SNR of the captured images.

Our system can also benefit the feature tracking task by improving image quality. We add an intensity camera, which is the same as we use in the prototype system, and set ``Auto Exposure (AE)'' as the exposure strategy. To evaluate the tracking performance, we introduce the Kanade–Lucas–Tomasi (KLT) feature tracker \cite{lucas1981iterative,tomasi1991detection} and record the numbers of the tracked features between consecutive frames. The more feature points tracked, the better the image quality. As shown in \cref{fig:klt,tab:klt}, images collected with AE strategy become blurry in the dynamic scene, which can barely provide any information for the feature tracking. In contrast, our prototype system, powered by our \myfore$_\textit{g}$ strategy, can produce sharper images, allowing successful KLT feature tracking. The feature tracking history displayed in \cref{fig:klt}(a) illustrates that our event-assisted imaging system can exhibit improved resistance to noise interference and enhanced stability after further incorporating the proposed \myback. Statistical analysis results shown in \cref{tab:klt} indicate that the proposed image acquisition strategy \myfore\ along with our image enhancement framework \myback can efficiently improve feature continuous tracking performance against the dynamic scenes.

\subsubsection{Robustness under Various Conditions} 
\begin{figure}[]
    \centering
    \includegraphics[width=0.9\linewidth]{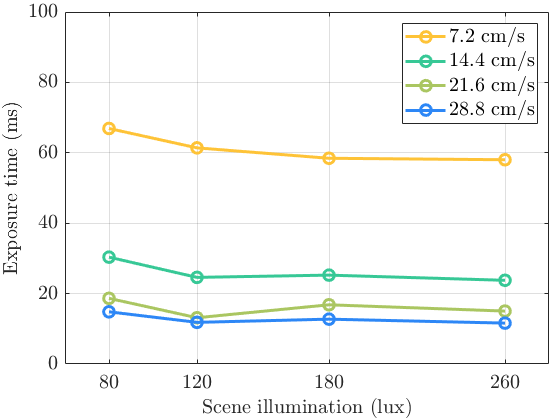}
    \caption{Average exposure times under various illuminations when shooting a static scene by the moving prototype system with various velocities.}
    \label{fig:illumination_speed}
\end{figure}

To assess the robustness of the proposed \myfore\ system, we conduct an experiment using our prototype system to capture a static scene under different illuminations and measure the average exposure times. The prototype system is installed on a programmable sliding trail to maintain uniform linear motion. As shown in \cref{fig:illumination_speed}, our prototype system with \myfore$_\textit{g}$ can sense the speed of scene movement through the event camera and stably ensure no motion blur, which is not seriously affected by illumination changes.

\subsection{Efficiency of Real-time Prototype Implementation}
\label{sec:exp_time}
Our proposed \myfore\ prototype is implemented by Python and operated on an RK3588S@2.40 GHz processor of the Orange Pi 5. The event capturing and readout latency is about 200 $\mu s$ as described in the white paper provided by iniVation Group \cite{inivation2020understanding}. According to the test, the average latency of thresholding and switching trigger level in the development board is 170 $\mu s$, while the response time of the intensity camera to external trigger signals is about 140 $\mu s$. Thus, the total latency of the prototype system is less than 1 $ms$, which can be considered a real-time system because the exposure time of an image is generally in the order of 10 $ms$.

\section{Limitations and Future Works}
Assuming that events are triggered only by motion, the proposed \myfore+\myback system is subject to events from other sources, \eg, illumination change and light flicker. In the future, we will exploit methods like event deflicker \cite{wang2022linear} to address this sensing limitation.

\section{Conclusion}
This paper provides a novel scene-adaptive exposure imaging strategy for intensity cameras, consisting of a Neuromorphic Shutter Control (\myfore) system and a Self-supervised Event-based Image Denoising (\myback) framework. To realize the real-time control of the camera shutter, we take advantage of the high temporal resolution of the event camera to monitor the intra-frame motion information and propose multiple efficient Event-based Motion Measure (EMM) functions to avoid motion blurs. For \myback, we propose to consider the statistics of noises and motion recorded by events for the generation of reliable artificial targets. To demonstrate the effectiveness and robustness of our event-based non-uniform imaging framework, we implement it in hardware and collect a real-world dataset, \ie, \mydata, containing the frames captured with \myfore\ and the corresponding events to facilitate future research in real-world scenarios.

\bibliographystyle{IEEEtran}
\bibliography{main.bib}

\vfill

\end{document}